\documentclass[10pt]{article}

\usepackage{authblk}
\usepackage{natbib}
\usepackage{mathtools}
\usepackage{amsmath}
\usepackage{amsthm}
\usepackage{amssymb}
\usepackage{amsbsy}
\usepackage{amsfonts}
\usepackage{amscd}
\usepackage{mathrsfs}
\usepackage{bm}
\usepackage{dsfont}
\usepackage{breqn}
\usepackage{color, soul}
\usepackage{enumerate}
\usepackage{empheq}
\usepackage{rotating}
\usepackage{ftnxtra}
\usepackage{fnpos}
\usepackage[titletoc,toc,title]{appendix}
\usepackage{euscript}
\usepackage{graphicx}
\usepackage{epsfig}
\usepackage{epstopdf}
\usepackage[lined,linesnumbered,ruled,commentsnumbered]{algorithm2e}
\SetKwProg{Init}{Initialization:}{}{}
\DeclareGraphicsExtensions{.pdf,.png,.jpg,.eps}
\usepackage{pstool}
\usepackage{upgreek}
\usepackage{mathrsfs}
\usepackage[inkscapelatex=false]{svg}

\usepackage{psfrag}

\usepackage{booktabs}

\numberwithin{equation}{section}

\theoremstyle{plain}	
\newtheorem{thm}{Theorem}[section]

\newtheorem*{prop*}{Proposition}
\theoremstyle{definition}	

\newtheorem{remark}[thm]{Remark}

\usepackage{caption}
\usepackage{subcaption}

\usepackage{hyperref}
\hypersetup{colorlinks=true, linkcolor=blue}
\hypersetup{colorlinks=true,citecolor=blue}

\DeclareMathAlphabet{\mathpzc}{OT1}{pzc}{m}{it}

\usepackage{amsmath, amsthm, amssymb}

\usepackage{cleveref}

\DeclarePairedDelimiter\abs{\lvert}{\rvert}

\DeclareMathOperator*{\maxx}{argmax}

\DeclareMathOperator*{\argminn}{argmin}

\makeatletter
\newsavebox{\@brx}
\newcommand{\llangle}[1][]{\savebox{\@brx}{\(\m@th{#1\langle}\)}%
  \mathopen{\copy\@brx\mkern2mu\kern-0.9\wd\@brx\usebox{\@brx}}}
\newcommand{\rrangle}[1][]{\savebox{\@brx}{\(\m@th{#1\rangle}\)}%
  \mathclose{\copy\@brx\mkern2mu\kern-0.9\wd\@brx\usebox{\@brx}}}%
\let\oldabs\abs
\def\abs{\@ifstar{\oldabs}{\oldabs*}}
\makeatother

\usepackage{accents}

    {\end{bmatrix}}%

\begin{document}

\title{\textsc{Mb}\textsc{Explainer}: \textbf{Multilevel bandit-based explanations for downstream models with augmented graph embeddings}}

\author[]{Ashkan Golgoon\thanks{ Corresponding author; \texttt{ashkangolgoon@gmail.com}}}
\author[]{Ryan Franks\thanks{\texttt{ryanfranks1996@gmail.com}}}
\author[]{Khashayar Filom\thanks{ \texttt{khashayar.1367@gmail.com}}}
\author[]{Arjun Ravi Kannan\thanks{\texttt{arjun.kannan@gmail.com}}}
\affil[]{\small \textit{Emerging Capabilities Research Group, Discover Financial Services Inc., Riverwoods, IL}}

\maketitle

\begin{abstract} 
Graph Neural Networks (GNNs) are a highly useful tool for performing various machine learning prediction tasks on graph-structured data. GNNs are widely used in the industry for a variety of applications. In many industrial applications, it is common that the graph embeddings generated from training GNNs are used in an ensemble model where the embeddings are combined with other tabular features (e.g., original node or edge features) in a downstream machine learning task. The tabular features may even arise naturally if, e.g., one tries to build a graph such that some of the node or edge features are stored in a tabular format. The goal of this paper is to address the problem of explaining the output of such ensemble models for which the input features consist of learned neural graph embeddings (possibly combined with additional tabular features). Therefore, we propose \textsc{Mb}\textsc{Explainer}, a model-agnostic explanation approach for downstream models with augmented graph embeddings.
\textsc{Mb}\textsc{Explainer} returns a human-comprehensible triple as an explanation (for an instance prediction of the whole pipeline) consisting of three components: a subgraph with the highest importance, the topmost important nodal features, and the topmost important augmented downstream features. A game-theoretic formulation is used to take the contributions of each component and their interactions into account by assigning three Shapley values corresponding to their own specific games. Finding the explanation requires an efficient search through the local search spaces corresponding to each component. \textsc{Mb}\textsc{Explainer} applies a novel multilevel search algorithm that enables simultaneous pruning of local search spaces in a computationally tractable way. In particular, three interweaved Monte Carlo Tree Search are utilized to iteratively prune the local search spaces. \textsc{Mb}\textsc{Explainer} also includes a global search algorithm that uses contextual bandits to efficiently allocate pruning budget among the local search spaces.  
We demonstrate the effectiveness of \textsc{Mb}\textsc{Explainer} by presenting a series of comprehensive numerical examples on multiple public graph datasets for both node and graph classification tasks.
\end{abstract}

\begin{description}
\item[Keywords:] Cooperative Game Theory, GNNs, Graph Explainability, Machine Learning Interpretability, Monte Carlo Sampling, Monte Carlo Tree Search, Shapley Values.
\end{description}

\tableofcontents


\section{Introduction}
\label{intro}

Geometric data are now ubiquitous in our day-to-day life from social networks to networks of neurons and molecular structures. Due to their generalized message passing scheme Graph Neural Networks (GNNs) offer a very powerful framework for performing representation learning on geometric data. The goal of a neural graph embedding method is to learn some lower dimensional representation of the graph structure. Deep graph models have been successfully used in various fields such as modeling financial transactions \citep{khazane2019deeptrax,van2022inductive_44}, Physics \citep{shlomi2020graph,eliasof2021pde}, drug discovery \citep{jiang2021could,gaudelet2021utilizing}, and recommender systems \citep{wu2022graph} (see \citep{zhou2020graph}). Improving the performance of GNNs has been under intense investigations in recent years. Some of these works include graph convolution \citep{kipf2016semi_1,hamilton2017inductive_2,li2020deepergcn_6}, graph attention \citep{velivckovic2017graph_3,wang2019heterogeneous}, graph transformers \citep{hu2020heterogeneous,shi2020masked}, and graph pooling \citep{yuan2020structpool}. 


Deep graph models, despite their robust capabilities, fall short in terms of transparency due to their black-box nature \citep{yuan2021explainability_13}. This makes it hard to generate human-comprehensible explanations for GNNs predictions \citep{ying2019gnnexplainer_22}. Nevertheless, it is important to generate explanations for GNNs or downstream models that utilize neural graph embeddings. This is because in many decision-critical applications (e.g., healthcare or financial services) it is vital (if not required by regulations)\footnote{For instance, explainability is required to generate Adverse Action Reason Codes (AARC) for use cases where graph embeddings are leveraged for credit decisioning.} to explain the predictions of deep graph models. Even for non-decision-critical fields, explainablity is still useful because it can generate insights about the graph structure by identifying the most important components of the input data that render a given prediction (see, e.g., \citep{wong2023discovery_24}). 

Generating explanations for GNNs has been the focus of extensive research during past years. Some of the notable contributions include: GNNExplainer \citep{ying2019gnnexplainer_22}, SubgraphX \citep{yuan2021explainability_13}, PGExplainer \citep{luo2020parameterized_23}, PGM-Explainer \citep{vu2020pgm}, and CF-GNNExplainer \citep{lucic2022cf} (see \S\ref{existing} for a comprehensive review of the existing GNNs explainability methods). These methods can be used to effectively explain predictions of GNNs.

In many practical applications, however, learned graph embeddings (from an upstream GNNs model) are in some form or shape utilized (often in conjunction with extra tabular features) in order to train a downstream machine learning model for a particular task. Note that the tabular features may be identical to node or edge features of the underlying graph in case the graph was built using some tabular data. In these situations, it is practical to leverage additional tabular machine learning models (e.g., GBDTs) to be able to natively encode the features, especially when categorical features and missing values are present (see, \S\ref{trstr} for more details). In this paper, we propose \textsc{Mb}\textsc{Explainer} a multilevel bandit-based explainability approach for downstream models with augmented graph embeddings. The method offers an easy way to understand explanations for the ensembles models where graph neural embeddings are utilized. An explanation generated by our devised method encodes the intermediate neural embeddings into an easy-to-interpret subgraph structure (and its node features) with the highest importance as well as the topmost important features from a downstream model. Therefore, \textsc{Mb}\textsc{Explainer} can generate an explanation for the whole pipeline composed of: upstream graph embeddings with a complex geometric structure and augmented tabular features in downstream with no geometric structure. In particular, \textsc{Mb}\textsc{Explainer} returns the subgraph, its node features, and downstream features as the most important components of the pipeline responsible for the prediction.



\textsc{Mb}\textsc{Explainer} applies an advanced pruning strategy that can provide an efficient way of searching in multiple local search spaces simultaneously. Specifically, local search spaces consisting of space of subgraphs, their node features, and augmented downstream features are pruned using three interweaved Monte Carlo Tree Search. Moreover, \textsc{Mb}\textsc{Explainer} implements a global search strategy using a contextual bandits algorithm that efficiently allocates pruning budget among local search spaces to enable simultaneous pruning of the spaces during the course of finding a desired explanation.



This paper is structured as follows. In \S\ref{prel}, we tersely review the fundamentals of graph neural networks as well as existing explanation methods for deep graph models. Some examples of downstream models with neural graph embeddings and their training pipelines are given in \S\ref{trstr}. In \S\ref{dwstream}, we lay the problem of explaining neural graph embeddings in downstream machine learning tasks with the formal formulation of \textsc{Mb}\textsc{Explainer} given in \S\ref{exdws}. Acceleration strategies to make the algorithms developed in \S\ref{dwstream} tractable are given in \S\ref{acc}. Numerical experiments on public datasets are given in \S\ref{numerics}. We use various examples to demonstrate the efficiency of \textsc{Mb}\textsc{Explainer} on public datasets. In particular, we report the experimental results for graph classification tasks on the MUTAG, PROTEINS, and Binarized ZINC as well as the results for the ogbn-arxiv dataset for node classification tasks.

\section{Related Work}
\label{prel}

In this section, we review some fundamental concepts and formulations of graph neural networks (see, \citep{kipf2016semi_1,hamilton2017inductive_2,velivckovic2017graph_3,hu2019strategies_4,wang2019dynamic_5,li2020deepergcn_6} for more details).

Let $\mathcal{G}=(\mathcal{V},\mathcal{E})$ denote a graph with vertices $\mathcal{V}=\{v_1,v_2,\cdots,v_N\}$ and edges $\mathcal{E}\subseteq\mathcal{V}\times\mathcal{V}$. An edge $e_{ij}=(v_i,v_j)\in \mathcal{E}$, connects vertices $v_i$ and $v_j$ if the graph is undirected.\footnote{For a directed graph, $e_{ij}$ is a directed edge that only allows messages passing from the source node $e_i$ to the target node $e_j$.} We denote the features of a node $v$ and an edge $e$ by $\mathbf{h}_v\in\mathbb{R}^d$ and $\mathbf{h}_e\in \mathbb{R}^b$, respectively.

\paragraph{Graph Neural Networks formulation.}
In this section, we discuss some background on graph representation learning, including message passing and aggregation schemes \citep{fey2019fast_7,yuan2021explainability_13, fang2023cooperative_14}. For simplicity, we assume that only node features are updated at each layer \citep{li2020deepergcn_6}.\footnote{Note that one can potentially obtain hidden representations for edges instead of nodes, or for both nodes and edges at the same network, but with two different types of weights.} Therefore, for a node $i$ and its neighborhood $\mathcal{N}(i)$, one defines
\begin{equation}
    \mathbf{m}^k_{ij}=\phi^k\left(\mathbf{h}^k_i, \mathbf{h}^k_j, \mathbf{h}^k_{e_{ij}}\right),\quad j\in\mathcal{N}(i),
\end{equation}
\begin{equation}
   \mathbf{m}^k_i=\gamma^k\left(\left\{\mathbf{m}^k_{ij}\,|\,j\in\mathcal{N}(i)\right\}\right),\qquad\quad\,
\end{equation}
\begin{equation}
    \mathbf{h}^{k+1}_i=\eta^k\left(\mathbf{h}^k_i,\mathbf{m}^k_i\right),\qquad\qquad\qquad\qquad
\end{equation}
where $\phi^k$, $\gamma^k$, and $\eta^k$ are differentiable functions that, respectively, perform message creation, message aggregation, and latent node representation update. Note that we require that the message aggregation function $\gamma$ be permutation invariant with respect to an arbitrary node ordering of the neighborhood $\mathcal{N}(i)$ for a given node $i$.


\paragraph{Summary of Machine Learning Tasks on Graphs.}
There are several types of machine learning prediction tasks on graph (geometric) data, namely:
\begin{itemize}
    \item Node classification
    \item Link prediction
    \item Community detection
    \item Graph classification (network similarity)
    \item Subgraph generation (generative methods)
\end{itemize}
Examples for node classification tasks include categorizing users/items into different classes/labels where users/items are represented as nodes in a graph. Link prediction examples include product recommendations or reward personalization where different products (or reward categories) are connected to different users and we are interested in finding potential connections between customers and new products. Community detection can be used to find densely linked clusters of nodes, e.g., finding a social circle or behavioral properties in different segments of the customer population. Graph classification tasks include protein discovery or drug side effects predictions. Subgraph generation can be used to generate novel graphs with desired properties leveraging methods like GANs. Our pipeline for interpreting downstream models with neural graph embeddings can be applied to all the machine learning tasks on graphs mentioned above. 

\begin{remark}[the definition of the computational graph for each ML task]
    Note that the computational graph for the above machine learning tasks are as follows:
    \begin{itemize}
        \item Node classification: $k$-hop neighborhood of a target node.
        \item Link prediction: $k$-hop neighborhoods of the starting and ending nodes of a link/edge.
        \item Community detection: $k$-hop neighborhoods of community nodes.
        \item Graph classification and subgraph generation: $k$-hop neighborhoods of the nodes in the subgraph.
    \end{itemize}
\end{remark}
\subsection{Review of the existing interpretability methods for GNNs}
\label{existing}
Next, we review some of the commonly used graph explanation methods from the literature. For the rest of this section, we generally follow the taxonomy provided by \citet{yuan2022explainability} in a recent survey. One may classify explanation methods for GNNs into five broad categories: gradient, perturbation, decomposition, and surrogate methods for instance-level explanations, and generative methods for model-level explanations.

Methods based on perturbing a continuous (or a discrete) mask have been intensely studied by GNNs explainability researchers \citep{ying2019gnnexplainer_22,luo2020parameterized_23,yuan2021explainability_13,schlichtkrull2020interpreting}. These methods investigate the effects of input perturbations on how it varies the output predictions of deep graph models \citep{yuan2022explainability}.

One of the early approaches addressing explainability of GNNs (for instance-level explanations) was proposed by \citep{ying2019gnnexplainer_22}. Here, we summarize the idea of the GNNExplainer method. Let us consider the node classification task for a node $v$ for which the predicted class probability is given by $\hat{\mathbf y}$. The computational graph of node $v$ is given by $\mathcal{G}_c(v)=\left(\boldsymbol{\mathcal{A}}_c(v),\boldsymbol{\mathcal{X}}_c(v)\right)$, where $\boldsymbol{\mathcal{A}}_c(v)\in\{0,1\}^{n\times n}$ and $\boldsymbol{\mathcal{X}}_c(v)=\{\mathbf{x}_i|v_i\in\mathcal{G}_c(v)\}$, are, respectively, the corresponding adjacency matrix and the node feature set.\footnote{Note that in \citep{ying2019gnnexplainer_22} it is assumed that edges do not possess any features.} Introducing the continuous mask $\boldsymbol{\mathcal{M}}_E$ over the edges and the binary mask $\boldsymbol{\mathcal{M}}_F$ over the node feature set, GNNExplainer tries to limit the computational graph $\mathcal{G}_c(v)$ to a subgraph $\mathcal{G}_s(v)\subseteq\mathcal{G}_c(v)$ with subfeatures $\boldsymbol{\mathcal{X}}_s(v)$ having the maximum mutual information with the prediction $\hat{\mathbf y}$. More formally, the masks $\boldsymbol{\mathcal{M}}_E$ and $\boldsymbol{\mathcal{M}}_F$ are obtained by solving the following optimization problem
\begin{equation}
\label{mu_max_ex}
\maxx_{(\boldsymbol{\mathcal{M}}_E,\boldsymbol{\mathcal{M}}_F)}\bigg[\mathrm{MI} (Y, (\boldsymbol{\mathcal{M}}_E,\boldsymbol{\mathcal{M}}_F)) = \mathrm{H}(Y)-\mathrm{H}(Y|(\boldsymbol{\mathcal{A}}',\boldsymbol{\mathcal{X}}'))\bigg]\,,
\end{equation}
where $\boldsymbol{\mathcal{A}}'=\boldsymbol{\mathcal{A}}_c\,\odot\sigma(\boldsymbol{\mathcal{M}}_E)$ and $\boldsymbol{\mathcal{X}}'=\boldsymbol{\mathcal{X}}_c\,\odot\boldsymbol{\mathcal{M}}_F$.\footnote{One can think of $\boldsymbol{\mathcal{A}}'$ as a \emph{fractional} adjacency matrix for the subgraph $\mathcal{G}_s$, i.e., $\boldsymbol{\mathcal{A}}'\in[0,1]^{n\times n}$ subjected to $\boldsymbol{\mathcal{A}}'[i,j]\leq\boldsymbol{\mathcal{A}}_c[i,j]$, $\forall v_i, v_j\in\mathcal{G}_c$.}

Using a similar mutual information-based formulation as that of GNNExplainer, \citet{luo2020parameterized_23} suggested another approach known as PGExplainer for generating explanations for multiple instances of GNNs. PGExplainer utilizes a parameterized mask predictor in order to learn discrete edge masks. The PGExplainer works as follows \citep{yuan2022explainability} (see Fig. 2 in \citep{luo2020parameterized_23}): Corresponding node embeddings are concatenated to obtain an edge embedding for a given edge in the graph. Edge embeddings are used to train a predictor that generates importance scores for edges. Next, the reparameterization trick is used to sample a random subgraph from edge distributions, which is then fed to the GNN to obtain the updated prediction. The mask predictor is then trained to maximize the mutual information between the original and updated predictions.


Apart from their successful applications in generating explanations in the image processing domain, surrogate methods have also been utilized in explaining deep graph models \citep{yuan2022explainability}. Some of the notable surrogate methods for GNNs explanations include GraphLIME \citep{huang2022graphlime}, PGM-Explainer \citep{vu2020pgm}, and RelEx \citep{zhang2021relex}.

GraphLIME \citep{huang2022graphlime} is an extension of a locally interpretable algorithm known as LIME (for models trained on tabular data) to explain deep graph models. The method leverages the so-called Hilbert-Schmidt Independence Criterion (HSIC) Lasso \citep{yamada2014high} feature selection method to build a nonlinear surrogate model that locally fits the dataset. Therefore, the method can select important node features for a given node classification task. However, GraphLIME explanation does not include information about the structure of the graph, e.g., it cannot identify which nodes or edges are more important for the prediction. Hence, the method can be applied to node classification tasks and not directly to graph classification scenarios \citep{yuan2022explainability}.

There have been recent efforts in extending cooperative game-theoretic feature attribution methods (e.g., Shapley values \citep{vstrumbelj2014explaining,lundberg2017unified_9}), previously employed for explaining machine learning predictions on tabular data, to GNNs \citep{yuan2021explainability_13,duval2021graphsvx}. 

Attempting to unify existing graph explainers into a single framework, \citet{duval2021graphsvx} proposed a Shapley value-based explanation approach named as GraphSVX. This method introduces a model-agnostic instance-level GNNs explanation technique that constructs a perturbed dataset $\mathcal{D}=\{(\mathbf{z}, f(\mathbf{z}')) \,|\, \mathbf{z}=(\mathcal{M}_N\mathbin\Vert\mathcal{M}_F),\,\mathbf{z}'=(\boldsymbol{\mathcal{A}}',\boldsymbol{\mathcal{X}}')\}$, on which a surrogate model $g$ is built.\footnote{Note that $(\mathcal{M}_N,\mathcal{M}_F)$ denotes the binary masks for nodes and features. The adjacency matrix and the feature matrix are, respectively, given by $(\boldsymbol{\mathcal{A}}',\boldsymbol{\mathcal{X}}')$.} Utilizing ideas from Shapley kernel construction (see \citep{lundberg2017unified_9}), the explanation $\phi$ of $f$ is obtained by solving the following optimization problem \citep[Eq. 3]{duval2021graphsvx}:
\begin{equation}
    \phi = \argminn_{g\in\Omega} \mathcal{L}_{f,\pi}(g)\,,
\end{equation}
where
\begin{equation}
\mathcal{L}_{f,\pi}(g)=\sum_{\mathbf z}\left(g(\mathbf z)-f(\mathbf{z}')\right)^2\pi_{\mathbf z}\,,
\end{equation}
and the weights $\pi$ are given by
\begin{equation}
    \pi_{\mathbf z}=\frac{F+N-1}{(F+N)\cdot |\mathbf z|}\cdot\binom{F+N-1}{|\mathbf z|}^{-1}\,.
\end{equation}
It turns out that the coefficients $\phi$ of the explanation model $g$ that GraphSVX computes are in fact an extension of Shapley values to graphs where the players consist of nodes and features. The learned parameters for $g$ are returned at the end as the explanation by GraphSVX.

SubgraphX \citep{yuan2021explainability_13} is another Shapley value-based explanation method. The method leverages the Monte Carlo Tree Search (MCTS) algorithm \citep{silver2017mastering_15} and Shapley-values as the reward function to find the most important subgraphs for a given instance of GNNs prediction. In particular, SubgraphX utilizes a node pruning action in the MCTS which one can think of as a mask generation algorithm applied to different subgraphs in the computaitonal graph \citep{yuan2022explainability}. 

Decomposition methods are another family of GNNs explainability approaches. Adapting the idea from computer vision \citep{montavon2019layer}, these methods try to map the prediction score to the input space. \citet{schnake2021higher} proposed an approach named GNN-LRP that extends Layer-wise Relevance Propagation (LRP) originally developed for explaining (convolutional neural networks) CNNs to graph-structured data. GNN-LRP works by propagating relevance backward through GNN layers, redistributing the model's output across walks within the graph structure. For each layer, it calculates relevance scores for individual nodes and edges, and for combinations of connected edges (i.e., walks). Long-range dependencies and relationships between nodes are captured using higher-order Taylor expansions. The GNN-LRP method is adaptable to various GNN architectures and has proven effective in various domains like analyzing structure-property relationships in quantum chemistry.

Instance-level explanation methods for deep graph models have been thoroughly studied in the literature. However, there are circumstances in which one tries to gain insights about GNNs predictions at a high-level going beyond an instance \citep{yuan2022explainability}. This is where model-level explanation methods play a pivotal role. 

XGNN \citep{yuan2020xgnn} proposes a reinforcement learning-based pipeline for explaining GNNs at the model-level by generating interpretable graph patterns. It starts with a graph generator that constructs graphs node by node, optimized via a policy gradient method to maximize the GNN’s confidence for a target class. A reward function guides the generator based on the model's output. XGNN follows specific structural constraints to ensure that generated graphs are valid according to some graph rules, e.g., by avoiding disconnected graphs. Moreover, XGNN promotes pattern diversity through regularization, helping to reveal human-interpretable structures that can provide meaningful model-level explanations.

\section{Downstream Neural Graph Embeddings}
\label{trstr}

In this section, we provide some insights about building downstream machine learning models using neural graph embeddings. The purpose of this section is to provide the reader with scenarios that building downstream models with neural graph embeddings are useful. In particular, we discuss efficient end-to-end training of both GNNs and the downstream models as well as strategies for categorical features encoding. Note that the interpretability (or explainability) module of \textsc{Mb}\textsc{Explainer} and its underlying algorithms (developed in Section \ref{dwstream}) are agnostic to how the GNNs model and the downstream model are trained. In fact, \textsc{Mb}\textsc{Explainer} only assumes that the trained GNNs $f^e$ and the downstream model $f^d$ are supplied to the algorithm.

\subsection{End-To-End Formulation and Pipeline Training}
GNNs are very efficient in encoding the underlying geometric structure between data points. However, they often have suboptimal performance when it comes to non-geometric data like tabular data (especially, when heterogeneous \citep{ivanov2021boost_39}). On the other hand, Gradient Boosted Decision Trees (GBDTs) perform very well on tabular data. Therefore, we propose novel architectures and training strategies for leveraging the capabilities of both GNNs and GBDTs when graph embeddings are augmented with tabular features in a downstream machine learning model. In particular, we discuss employing GBDTs as either a stand-alone or iterative encoder (and imputer) to enrich node (or edge) features of GNNs. Additionally, we discuss concurrent (end-to-end) training of GNNs and downstream machine learning models (either a GBDT or a general model architecture). Our primary intention for devising a concurrent training strategy is to make use of the backpropagation errors from both models to concurrently update both models’ parameters in order to achieve a more optimally trained modeling pipeline with faster convergence overall.

\subsubsection{Leveraging GBDTs to natively handle categorical features and missing values}

One caveat of utilizing DNNs in general—and GNNs in particular—is that they are simply unable to handle and encode categorical features and missing values natively. However, it is possible to employ tree-based models, along with GNNs to leverage their superior native encoding and imputing mechanisms (see, e.g., \citep{ivanov2021boost_39, stekhoven2012missforest_46}).  

\paragraph{Stand-alone GBDTs as encoder.}

\begin{figure}[!htb]
	\centering
    \centerline{\includesvg[width=0.85\columnwidth]{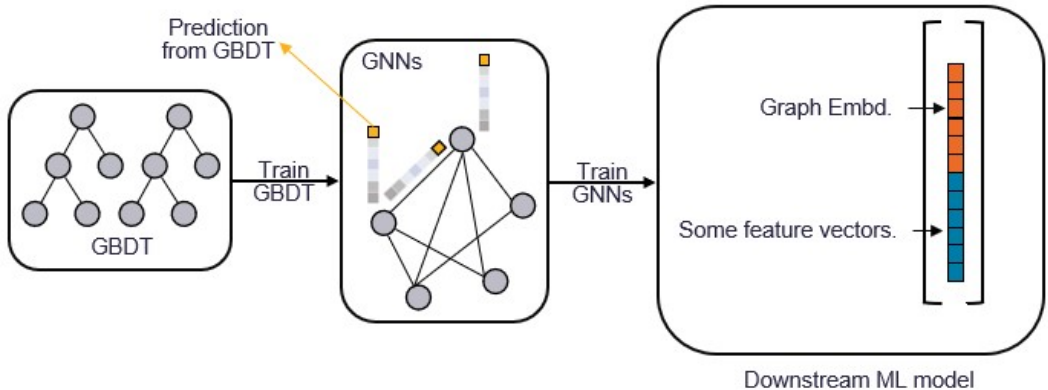}}
	\caption{Stand-alone GBDT employed as an encoder to generate additional features for graph nodes or graph edges. The obtained neural embeddings are augmented with additional feature in a downstream ML model.}
	\label{fig_1}
\end{figure}

In obtaining the graph neural embeddings from the GNNs model, we propose to train a GBDT using node (or edge) features on a supervised task, e.g., node classification or link prediction. Concretely, we utilize the output predictions (probability scores) from the trained GBDT as an extra feature that is added to the graph either as a node (or an edge) feature. Next, the neural graph embeddings are obtained by training the GNNs model. This way, we are able to take advantage of superior categorical features and missing values encoding handled efficiently by a GBDT approach, e.g., a CatBoost model. Finally, similar to the previous setup, the obtained embeddings are used, along with additional features, in training a downstream model. This process is schematically shown in Figure \ref{fig_1}.

\paragraph{Iterative encoder using GBDTs.}

\begin{figure}[!htb]
	\centering
    \centerline{\includesvg[width=0.85\columnwidth]{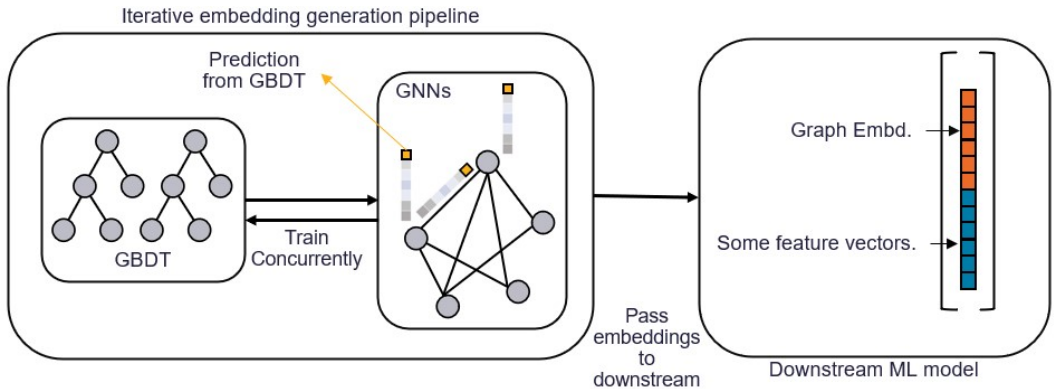}}
	\caption{Starting with a GBDT, neural graph embeddings are iteratively optimized by training both the GBDT and GNNs concurrently. The obtained neural embeddings are augmented with additional feature in a downstream ML model.}
	\label{fig_2}
\end{figure}

The previous architecture can be set up such that we can update the GBDTs and GNNs model parameters iteratively so that both models can be trained simultaneously (see, e.g. \citep{ivanov2021boost_39}). The generated embeddings will again be used in a downstream model for training (see Figure \ref{fig_2}). Let us consider the computational graph of a target node $v$, i.e., $\mathcal{G}_v (\boldsymbol{\mathcal{X}}_v, \boldsymbol{\mathcal{A}}_v)$, where $\boldsymbol{\mathcal{X}}_v$ and $\boldsymbol{\mathcal{A}}_v$ are, respectively, node features and the adjacency matrix of the target node neighborhood $\mathcal{N}(v)$. Tabular input features for GBDT model $f^t$ are given by $\boldsymbol{X}$ with their target labels $Y$. In each iteration we train $m$ (weak learners) trees from GBDT, pass their predictions as extra node features, i.e., $\mathbf{X}'$ to $\mathcal{G}_v$, train $n$ steps of GNNs model $f^e$ by updating both GNNs parameters and augmented feature from GBDT (i.e., $\mathbf{X}'$) using stochastic gradient descent (SGD), and update the target label for GBDT for the next iteration. The details of the algorithm used to iteratively generate embeddings are highlighted in Algorithm \ref{algo_1} (see also \citep{ivanov2021boost_39}). After this training process, we obtain the trained GNNs model and associated embeddings.

\begin{algorithm}
    \SetAlgoLined 
    \KwIn{Number of iterations $\kappa$, tabular node features for GBDT $\boldsymbol{X}$, computational graph of a target node $\mathcal{G}_v (\boldsymbol{\mathcal{X}}_v, \boldsymbol{\mathcal{A}}_v)$, target label $Y$ (initialize GBDT target $y=Y$) \citep{ivanov2021boost_39}.}
    \BlankLine
    \For{$i=1$ to $\kappa$}
    {
        \# Train $m$ trees of GBDT\\
        $f^t_i\leftarrow \argminn_{f^t_i} L_{GBDT}\left(f^t_i(\boldsymbol{X}),y\right)$, build $m$ trees of weak learners each time\\
        $f^t\leftarrow f^t_i+f^t$;\\
        $\boldsymbol{X}'=f^t(\boldsymbol{X})$;\\
        \# Train $n$ steps of GNNs on augmented node/edge features from GBDT\\
        $f^e, \boldsymbol{X}'\leftarrow \argminn_{f^e, \boldsymbol{X}'}L_{GNNs}\left(f^e_v\left(\mathcal{G}_v\left(\boldsymbol{\mathcal{X}}_v||\boldsymbol{X}', \boldsymbol{\mathcal{A}}_v\right)\right), Y\right)$, where we use SGD by finding the gradient of GNNs loss function w.r.t. model parameters of $f^e$ and augmented features $\boldsymbol{X}'$.\\
        \# Update the target for GBDT\\
        $y=\boldsymbol{X}'-f^t(\boldsymbol{X})$;
    }
    \KwOut{$f^e$, and thus, the corresponding embeddings to be passed to downstream model.}
\caption{ITERATIVE EMBEDDINGS GENERATION PIPELINE}\label{algo_1}
\end{algorithm}

\subsubsection{Concurrent training of graph embeddings and downstream models}

\begin{figure}[!htb]
	\centering
    \centerline{\includesvg[width=0.8\columnwidth]{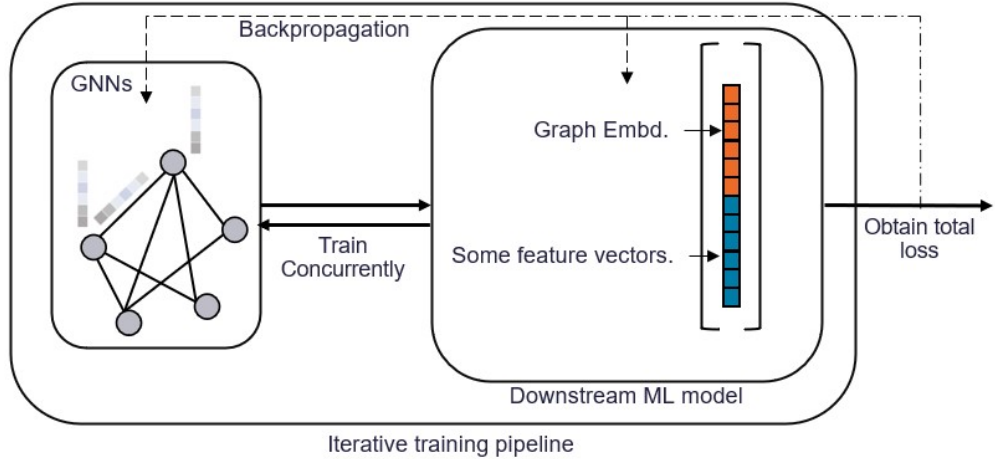}}
	\caption{Concurrent training of both GNNs and downstream models.}
	\label{fig_3}
\end{figure}

It is also possible to design a training pipeline such that both the GNNs and downstream models are trained concurrently. The details of the training algorithm depend ultimately on the type of a downstream model. For example, whether the model architecture allows for the model parameters update using SGD (or mini-batch SGD) like DNNs, or, whether the downstream model is a GBDT like CatBoost, where we can optimize the model iteratively using an approach similar to Algorithm \ref{algo_1} (i.e., by iteratively updating the target label for weak learners). Figure \ref{fig_3} illustrates our proposed training procedure. The details of the algorithm are outlined in Algorithm \ref{algo_2}. One needs to modify the algorithm given a particular model architecture of the downstream model.

\begin{algorithm}
    \SetAlgoLined 
    \KwIn{Number of iterations $\kappa$, tabular downstream features $\boldsymbol{x}_v$, computational graph of a target node $\mathcal{G}_v (\boldsymbol{\mathcal{X}}_v, \boldsymbol{\mathcal{A}}_v)$, target label $Y$.}
    \BlankLine
    \For{$i=1$ to $\kappa$}
    {
        \# Train $n$ steps of GNNs\\
        $f^e_i\leftarrow \argminn_{f^e} L_{GNNs}\left(f^e_v\left(\mathcal{G}_v \left(\boldsymbol{\mathcal{X}}_v, \boldsymbol{\mathcal{A}}_v\right)\right),Y\right)$, where we use SGD by finding the gradient of GNNs loss function w.r.t. model parameters of $f^e$.\\
        $\boldsymbol{h}^l_v=f^e_v(\mathcal{G}_v)$;\\
        \# Train $m$ steps of downstream ML model\\
        $f^d_i\leftarrow\argminn_{f^d_i}L_{DWST}\left(f^d\left(\boldsymbol{x}_v||\boldsymbol{h}^l_v\right),Y\right)$, use an optimizer that $f^d$ allows whether SGD for DNNs or through creating additional weak learners recursively for GBDTs.\\
        \# Update the GNNs model parameters $\theta$ by backpropagating the DWST loss to the GNNs model, i.e.,\\
        $\theta\leftarrow\theta-\eta\cdot\frac{\partial L_{DWST}}{\partial\boldsymbol{h}^l_v}\cdot\frac{\partial f^e_v\left(\mathcal{G}_v\left(\boldsymbol{\mathcal{X}}_v, \boldsymbol{\mathcal{A}}_v\right)\right)}{\partial\theta}$;
    }
    \KwOut{$f^e$, $f^d$;}
\caption{CONCURRENT TRAINING OF GNNS AND DOWNSTREAM ML MODELS }\label{algo_2}
\end{algorithm}


\begin{remark}[negative sampling]
Due to label-imbalanced ratio problem, GNN-based methods could have a poor performance on some supervised learning approaches where the node/edge label distribution is heavily skewed \citep{liu2021pick_41,zeng2019graphsaint_42,huang2022auc_43}. Our method allows a variety of negative sampling techniques to be used when computing graph embeddings. This includes simple seed sampling approaches where GNNs mini-batches frequencies are adjusted based on the imbalanced seed label ratio for training the model. Or, alternatively, one may choose more sophisticated negative sampling approaches such as Pick-n-Choose \citep{liu2021pick_41} and GraphSAINT \citep{zeng2019graphsaint_42}, where nodes neighboring structure are taken into account when constructing each training graph.
\end{remark}

\begin{remark}[handling categorical features through graph structure]
    In this work, we discussed how categorical features can be natively handled by leveraging a GBDT when training a GNNs in one pipeline. In some circumstances, it would be desirable to encode categorical features inside the graph structure. Depending on the cardinality of categorical features, one may create pseudo nodes (or artificial nodes), and thus, artificial links to create extra message passing paths in the graph by connecting nodes having identical categorical features (see, e.g., \citep{deutsch2024ms}). 
\end{remark}

\begin{remark}[handling dynamic/temporal or unstructured node/edge features]
    Temporal node features, e.g., time-series node features can be handled using a Recurrent Neural Networks model. Similarly, if we need to apply our method to graphs possessing unstructured node features, e.g., images, text, or voice data, we may leverage appropriate methods like CNNs, BERT (or Transformers, in general) to encode these features inside the pipeline. The model parameters associated with encoding node/edge features will be part of the computational graph (forward path) of the entire pipeline, and thus, their corresponding model wights will be updated concurrently with the rest of the pipeline model parameters.
\end{remark}

\begin{remark}[training GNNs and downstream models in different time periods]
    Note that in some cases it would be more desirable to train the downstream model and the GNNs using overlapping or non-overlapping time periods to reduce overfitting of both models. Finding the best splitting of the original training data to reduce overfitting requires conducting several training experiments with different splits of the training data. Moreover, one can also decide to train the two models with different sampling criteria in order to find the best sampling ratios at which both models would have a more superior performance.
\end{remark}

\section{MBExplainer}
\label{dwstream}

In this section, we carefully demonstrate the interpretability formulation for \textsc{Mb}\textsc{Explainer}. In particular, we formulate the interpretability problem for an ensemble model where neural graph embeddings generated from an upstream GNNs model are augmented with additional tabular features in a downstream ML model when performing predictions. To the best of our knowledge, this problem has not been addressed before in the literature. This novel interpretability approach offers a way to generate explanations for the whole pipeline which consists of two inherently different components: i) neural graph embeddings, possessing a complex geometric structure of the underlying graph data, and ii) augmented tabular data in downstream with absolutely no geometric structure. 

Note that one may be tempted to employ post-hoc explainers or self-interpretable explainers (if applicable) for the downstream model and treat the graph embeddings as some synthetic set of features.\footnote{As a matter of fact, treating graph embeddings as extra features and applying explainers to the downstream model would basically freeze the components of the computational graph and node features (and thus, the resulted embeddings $\mathbf{h}_v^k$)  in the GNN model $f^e$. Therefore, interactions among various subgraphs and their components and in connection with other downstream components (features) would be simply ignored.} However, this method fails to take into consideration the complex dependencies of graph embeddings and original features, and thus, leads to ambiguous explanations. Moreover, our devised interpretability approach takes all the interactions between downstream features and graph embeddings into consideration. Therefore, this approach results in an explanation that properly measures the true importance of each component. 

\subsection{Formulating Explanations at the Downstream Level}
\label{exdws}
In this section, we formulate the problem of graph embeddings explanations when performing predictions in a downstream model where in addition to the neural embeddings one uses augmented tabular features. Consider the previous setup, let $f^e$ and $f^d$, respectively, denote the trained GNNs model and the trained downstream model. Let the $k$-hop neighborhood of a target node $v$ be identified with $\mathcal{N}(v)$, consisting of the computational graph $\mathcal{G}_v$ with nodes $\left\{v_1,\cdots,v_m\right\}$ and the adjacency matrix $\boldsymbol{\mathcal{A}}_v \in\mathbb{R}^{m\times m}$. The node and edge features are, respectively, given by  $\boldsymbol{\mathcal{X}}_v\in\mathbb{R}^n$ and $\boldsymbol{\mathcal{X}}_{e_{uv}} \in\mathbb{R}^n$ vectors.\footnote{Here, for simplicity and without loss of generality, we make the assumption that node and edge features are of the same dimension.} The resulted graph embeddings for node $v$ from the $l$-th layer of the GNNs model are given by $\mathbf{h}_v^l=f_v^e (\mathcal{G}_v )\in\mathbb{R}^D$, where $D$ is the hidden dimension of the $l$-th output layer. The goal is to find an explanation for the whole pipeline viewed as a single model when the neural graph embeddings $\mathbf{h}_v^l$ are augmented with downstream features denoted by $\boldsymbol{x}_v\in\mathbb{R}^n$:\footnote{Without loss of generality, we assume that the downstream features and node features in the graph are of the same dimension. Note that often the downstream features are simply the same as tabular node features, i.e., $\boldsymbol{\mathcal{X}}_v=\boldsymbol{x}_v$. For instance, augmenting neural embeddings with original tabular node features in a downstream tree model (e.g., XGBoost) turns out to have a superior performance than making prediction solely based off the upstream GNNs model, especially when faced with heterogeneous data (see, e.g., \citep{ivanov2021boost_39,van2022inductive_44,van2023catchm_45}).}
\begin{equation}
\label{31}
    \boldsymbol{x}_v=\{x^v_1,\cdots,x^v_n\}\,.
\end{equation}
In other words, we are interested in generating an instance-level explanation for the prediction of the seed node $v$ when its embeddings $\mathbf{h}_v^l$ are used at downstream such that:
\begin{equation}
    \label{32}f^d\left(\boldsymbol{x}_v||\,\mathbf{h}^k_v\right)=f^d\left(\boldsymbol{x}_v||f^e_v\left(\mathcal{G}_v\right)\right)\,.
\end{equation}
For the rest of this paper, we implicitly assume that the explained node is $v$ and we drop the subscript for the sake of brevity.  

\paragraph{Game-Theoretical Explanation Formulation.}
Next, we propose to employ Shapley values (see, \citep{kuhn1953contributions_8,lundberg2017unified_9,chen2018shapley_10,2021arXiv210210878M_11}) to obtain various importance scores in the course of an explanation. In doing so, we introduce the triple $(S',\mathcal{G}',M)$ such that:
\begin{itemize}
    \item $S'$ denotes a downstream feature mask that is going to identify the topmost important augmented features at the downstream level.
    \item $\mathcal{G}'$ denotes a subgraph of the computational graph in the trained GNNs which is having the highest importance for the pipeline prediction of the seed node. 
    \item $M$ defines a node feature mask that is going to identify the topmost important nodal features in the subgraph $\mathcal{G}'$.\footnote{Note that one can similarly define an edge feature mask that identifies topmost important edge features in the subgraph $\mathcal{G}'$. Here, for the sake of brevity, we only work with a node feature mask, but the extension of the framework to include edge feature mask is trivial.} Note that, here, for the sake of simplicity, we assume that the feature mask is uniformly applied to all nodes. However, later we provide insights as to how the formulation changes if one wants to consider a nonuniform node feature mask. 
\end{itemize}
To quantify the contributions of each component, we assign three Shapley values each corresponding to their own specific game. Therefore,
\begin{equation}
\label{33}
\begin{split}
    \varphi^{(S',\mathcal{G}',M)}(S'):=&\mathlarger{\mathlarger\sum}_{S_1\subseteq N \setminus S'}\frac{|S_1|!.\left(n-s'-|S_1|\right)! }{\left(n-s'+1\right)!}.\bigg[f^d\left(x_{S_1}, x_{S'}, x^*_{N\setminus (S_1\cup S')}, f^e\left(\left(\mathcal{G}'\right)^M\right)\right)\\&-f^d\left(x_{S_1}, x^*_{N\setminus S_1}, f^e\left(\left(\mathcal{G}'\right)^M\right)\right)\bigg],
\end{split}
\end{equation}
\begin{equation}
\label{34}
\begin{split}
    \varphi^{(S',\mathcal{G}',M)}(\mathcal{G}'):=&\mathlarger{\mathlarger{\sum}}_{S_2\subseteq\{v_{k+1},\cdots,v_m\}}\frac{|S_2|!.\left(m-k-|S_2|\right)!}{\left(m-k+1\right)!}.\bigg[f^d\left(x_{S'}, x^*_{N\setminus S'}, f^e\left(\left(\mathcal{G}'\cup S_2\right)^M\right)\right)\\&-f^d\left(x_{S'}, x^*_{N\setminus S'}, f^e\left(\left(S_2\right)^M\right)\right)\bigg],
\end{split}
\end{equation}
\begin{equation}
\label{35}
\begin{split}
    \varphi^{(S',\mathcal{G}',M)}(M):=&\mathlarger{\mathlarger{\sum}}_{S_3\subseteq N\setminus M}\frac{|S_3|!.\left(n-|M|-|S_3|\right)!}{\left(n-|M|+1\right)!}.\bigg[f^d\left(x_{S'}, x^*_{N\setminus S'},f^e\left(\left(\mathcal{G}'\right)^{M\cup S_3}\right)\right)\\&-f^d\left(x_{S'}, x^*_{N\setminus S'},f^e\left(\left(\mathcal{G}'\right)^M\right)\right)\bigg],
\end{split}
\end{equation}
where the set of players for each game is defined as:
\begin{equation}
\label{36}
    P^{S'}:=\left\{S',\overbrace{x_{S'+1},\cdots,x_n}^{\text{features not in}\, S'}\right\},
\end{equation}
\begin{equation}
\label{37}
    P^{\mathcal{G}'}:=\left\{\left(\mathcal{G}'\right)^M,\overbrace{v_{k+1},\cdots,v_m}^{\text{nodes not in}\, 
 \mathcal{G}'}\right\},
\end{equation}
\begin{equation}
\label{38}
    P^M:=\left\{\overbrace{\mathcal{X}_1,\cdots,\mathcal{X}_{n-|M|}}^{\text{features not masked by}\, M},\overbrace{(\mathcal{X}_{n-|M|+1},\cdots,\mathcal{X}_n)}^{\text{features masked by}\, M}\right\}.
\end{equation}
In the first formula \eqref{33}, we iterate over subsets $S_1$ of augmented downstream features not masked by $S'$ ($s'$ denotes the size of $S'$). Thus, \eqref{33} gives the Shapley value of $S'$ in a game defined based on $(S',\mathcal{G}',M)$, whose players are given by $P^{S'}$, i.e., the mask $S'$ of the downstream features, along with downstream features not in $S'$. In the second formula \eqref{34}, we iterate over subsets $S_2$ of nodes which are not in $\mathcal{G}'$. Hence, \eqref{34} gives the Shapley value of $\mathcal{G}'$ in a game defined based on $(S',\mathcal{G}',M)$, whose players are given by $P^{\mathcal{G}' }$, i.e., the subgraph $\mathcal{G}'$  and the nodes of the computational graph $\mathcal{G}$ that are not in $\mathcal{G}'$ denoted by $\left\{v_{k+1},\cdots,v_m\right\}$. Finally, in \eqref{35} we iterate over subsets $S_3$ of node features not masked by $M$ (the size of $M$, which gives the number of features masked by $M$ is denoted by $|M|$). The players of a game defined with values in \eqref{35} are given by $P^M$ which includes the node features masked by $M$ (as a single player) and those not masked by $M$. Note that $x^*$ denotes a baseline datapoint from downstream feature distribution (often obtained by taking the average over the whole distribution). Moreover, we adopt a zero-padding strategy to compute terms such as $f^e (({\mathcal{G}'})^M)$ by setting features of nodes not belonging to $\mathcal{G}'$ to zero and applying the mask $M$ to features of nodes belonging to $M$.

Having defined the Shaley values for each component in $(S',\mathcal{G}',M)$, we define the following measurement to attribute an importance score to a triple $(S',\mathcal{G}',M)$ as a whole:
\begin{equation}
    \label{39}
    \varphi^{(S',\mathcal{G}',M)}(S',\mathcal{G}',M):=\varphi^{(S',\mathcal{G}',M)}(S')+\lambda^{\mathcal{G}'}.\varphi^{(S',\mathcal{G}',M)}(\mathcal{G}')+\lambda^M.\varphi^{(S',\mathcal{G}',M)}(M),
\end{equation}
where $\lambda^{\mathcal G'}$ and $\lambda^M$ are constants\footnote{Note that these constants are determined by expert knowledge and are possibly influenced by explanation requirements and strategies.} that control the relative importance of the subgraph $\mathcal{G}'$ and the node feature mask $M$ as components of the explanation with respect to each other as well as relative to the downstream feature mask $S'$. One can set these hyperparameters to fine-tune the explanation score or even use them to make handcrafted explanation measurement scores depending on different use cases. In Section \ref{acc}, we discuss acceleration strategies and algorithms utilized to implement the described explainability approach. A promising explanation needs to have a high $\varphi^{(S',\mathcal{G}',M)} (S',\mathcal{G}',M)$ value, and at the same time, it must be succinct enough so that the explanation is concise and useful.   
\begin{remark}[product set of players]
    One may be tempted to consider a game with a product set of players given by
    \begin{equation}
        \label{310}
        \left\{S', \overbrace{x_{S'+1},\cdots,x_n}^{\text{features not in}\, S'}\right\}\times \left\{(\mathcal{G}')^M,\overbrace{v_{k+1},\cdots,v_m}^{\text{nodes not in}\, \mathcal{G}'}\right\}.
    \end{equation}
    Note, however, that although \eqref{310} would consider all possible ways the downstream features and subgraph players can interact with, it comes at a significant price of increasing the computational complexity of obtaining Shapley values. Therefore, to lower the computational complexity, we work with players \eqref{36}, \eqref{37}, and \eqref{38} rather than \eqref{310}.
\end{remark}
\begin{remark}[uniform vs. non-uniform masking of graph node and edge features]
   One may ask what if one needs to learn a non-uniform mask for node (or edge) features. In that case the formulations (and corresponding games) in \eqref{36} – \eqref{38} can still be applied, but one needs to extend the feature mask dimension to include all the variations of masking for different nodes.  
\end{remark}

\begin{remark}[evaluation framework]

One can define many evaluation metrics to measure the effectiveness of an explanation. For example, we may employ the Fidelity score as one of the metrics that can give us some insights about the efficiency of an explanation. For a downstream model with neural graph embeddings, the Fidelity score is computed for $N_s$ total number of testing samples as follows:
\begin{equation}
    \label{311}
    \text{Fidelity}_{+}(S',\mathcal{G}',M):=\frac{1}{N_s}.\mathlarger{\sum}_{j=1}^{N_s}\left(\Big|f^d_{y_i}\left[x_N, f^e\left(\left(\mathcal{G}\right)^N\right)\right]-f^d_{y_i}\left[x^*_{S'},x_{N\setminus S'},f^e\left(\left(\mathcal{G}\setminus\mathcal{G}'\right)^{1-\mathds{1}(M)}\right)\right]\Big|\right),
\end{equation}
\begin{equation}
    \label{312}
    \text{Fidelity}_{-}(S',\mathcal{G}',M):=\frac{1}{N_s}.\mathlarger{\sum}_{j=1}^{N_s}\left(\Big|f^d_{y_i}\left[x_N, f^e\left(\left(\mathcal{G}\right)^N\right)\right]-f^d_{y_i}\left[x_{S'}, x^*_{N\setminus S'}, f^e\left(\left(\mathcal{G}'\right)^M\right)\right]\Big|\right),
\end{equation}
where $f_{y_i}^d$ gives the predicted probability of class $y_i$. Fidelity score evaluates the contribution of the generated explanation in forming the initial prediction by either occluding the explanation $(S',\mathcal{G}',M)$ from the original computational graph with no downstream and graph features masked yielding $\text{Fidelity}_{+}$ or by restricting the model to only use the explanation $(S',\mathcal{G}',M)$ yielding $\text{Fidelity}_{-}$ (see also \citep{fey2019fast_7,amara2022graphframex_19}). Note that although Fidelity measures how good an explanation is, it is not derived from a game-theoretic approach. A better way to evaluate an explanation is based on Shapley values which are used to find a good explanation in the first place as suggested in the paper. Therefore, we may use the computed Shapley value for the explanation, i.e., $\varphi^{(S',\mathcal{G}',M)} (S^',\mathcal{G}',M)$, as a score to evaluate the generated explanation $(S',\mathcal{G}',M)$.
\end{remark}
\begin{remark}[Evaluating the explanation in the presence of ground truth]
\label{re33}
Note that it is also possible to create synthetic benchmark datasets by attaching motifs to graphs that can be generated using a generative method. For example, we may utilize various graph generation algorithms such as Random Barabasi-Albert (BA) graphs \citep{albert2002statistical_37}, Random Erdos-Renyi (ER) graphs \citep{erdds1959random_38}, or grid graphs, along with randomly generating motifs (e.g., house motifs and cycle motifs) in order to create synthetic datasets with pre-determined ground-truth. Therefore, it is possible to generate a dataset with known ground truth for both geometric and non-geometric synthetic data and provide a comprehensive evaluation pipeline for the devised interpretability component of the pipeline.    
\end{remark}
\subsection{Acceleration Strategies and Implementation Architecture}
\label{acc}

In this section, we provide our acceleration strategies and algorithms that are developed and carefully tailored to make the computation of Shapley values, formulated in Section \ref{exdws}, tractable and scalable to large graphs and feature sets. Obtaining an explanation $(S',\mathcal{G}',M)$ through searching all possible combinations (a brute-force method) is either intractable or comes at a high computational expense. Therefore, here, we develop a novel search algorithm to overcome the computational barriers of finding a promising explanation. Particularly, we employ a multilevel bandit-based search strategy that iteratively prunes the search spaces of subgraphs, their node features, and downstream features simultaneously. The search strategy consists of two parts, namely the local and global search algorithms. The local search strategy includes pruning the space of subgraphs, their node features, and downstream features, whereas the global search algorithm allocates the pruning budget among the three local search spaces in an efficient manner. The local search spaces are navigated using three interweaved Monte Carlo Tree search. The global search strategy is developed using a contextual bandits type algorithm which drives efficient search among the local search spaces.

\subsubsection{Assembling the local search spaces}
\label{localsp}
We propose to utilize the Monte Carlo Tree Search (MCTS) (see, e.g., \citep{yuan2021explainability_13,silver2017mastering_15,jin2020multi_16}) in order to locally navigate the three search spaces, namely the space of subgraphs, node features, and augmented downstream features. Therefore, we employ three interweaved MCTSs that iteratively prune the local search spaces. Let us denote the search trees corresponding to the space of subgraphs, node features, and augmented downstream features by $\mathcal{T}^{\mathcal{G}' }$, $\mathcal{T}^M$, and $\mathcal{T}^{S' }$, respectively. Note that the tree root corresponding to $S'$ is given by $N$ (i.e., no augmented downstream features is set to the baseline $x^*$, and thus, $|S'|=0$) in the tree $\mathcal{T}^{S'}$. The root for $\mathcal{T}^{\mathcal{G}'}$ is the computational graph of the seed node $v$, i.e., $\mathcal{G}_v$ (with no nodes pruned).\footnote{Assuming that the task at hand is node classification.} Finally, the root for $\mathcal{T}^M$ corresponds to $N$ (i.e., no node features is masked, and thus, $|M|=0$). In the following, we describe the elements of the MCTS, including tree traversal (tree policy), node expansion, roll-out (Monte Carlo simulation or default policy), and back-propagation for the three pruning actions that iteratively limit the three local search spaces.

\paragraph{Assembling the MCTS for $\mathcal{T}^{S' }$, $\mathcal{T}^{G^' }$, and $\mathcal{T}^M$.}
In the following, we discuss how the MCTS algorithm navigates the search trees $\mathcal{T}^{\mathcal G' }$, $\mathcal{T}^M$, and $\mathcal{T}^{S' }$. Without loss of generality, we discuss the details of the MCTS for $\mathcal{T}^{\mathcal G' }$ as the process is similar for $\mathcal{T}^M$ and $\mathcal{T}^{S' }$. Let us consider the computational graph of the seed node $v$ in the GNNs model, i.e., $\mathcal{G}_v$. We are interested in pruning the computational graph to find a subgraph part of the explanation $(S',\mathcal{G}',M)$. Starting with the root node, corresponding to the computational graph $\mathcal{G}_v$, we build a search tree $\mathcal{T}^{G' }$ such that each node in the tree is a connected subgraph. A pruning action (i.e., an edge in the search tree) for $\mathcal{T}^{\mathcal{G}' }$ removes a node (and all edges connected with it) from its parent node containing the subgraph $\mathcal{G}'$. Here, the game (tree) state $s_i$ can be identified with a tree node $N_i$ corresponding to its subgraph $\mathcal{G}_i$. An action $a_j$ removes a node (and its corresponding edges) from $\mathcal{G}_i$ and yields the state $s_j$ corresponding to its node $N_j$ and the subgraph $\mathcal{G}_j$. The tree policy is given by (see, e.g., \citep{silver2017mastering_15,kocsis2006bandit_18,browne2012survey_20,CSE599i_21,yuan2021explainability_13}):
\begin{equation}
\label{41}
    a^*= \maxx_{a_j\in \mathcal{A}_i}\left[Q(s_i,a_j)+U(s_i,a_j)\right],
\end{equation}
where $\mathcal{A}_i$ is a set of available action at state $s_i$, and $Q(s_i,a_j )$ denotes the empirical average reward for multiple visits in the tree where action $a_j$ is selected from state $s_i$. The term $U(s_i,a_j )$ is a confidence interval bound type function that controls exploration versus exploitation when the agent lacks high confidence in estimating the state-action pair value. Similar to \citep{yuan2021explainability_13, silver2017mastering_15, CSE599i_21}, we define $U$ as 
\begin{equation}
\label{42}
    U(s_i, a_j)=c.P(s_i,a_j).\frac{\sqrt{\Sigma_{a_k\in\mathcal{A}_i}\textrm{N}(s_i,a_k)}}{1+\textrm{N}(s_i, a_j)},
\end{equation}
where $c$ is a hyperparameter that one can use to control the exploration-exploitation trade-off; $P(s_i,a_j )$ denotes the prior probability of selecting action $a_j$ from state $s_i$; and $\textrm{N}(s_i,a_j)$ is the number of visit counts for playing an action $a_j$ for a given state $s_i$. One may express the empirical average reward $Q(s_i,a_j)$ in terms of total rewards $w(s_i,a_j)$ for multiple visits $\textrm{N}(s_i,a_j)$ from the state-action pair $(s_i,a_j )$ as follows
\begin{equation}
\label{43}
    Q(s_i,a_j)=\frac{w(s_i,a_j)}{\textrm{N}(s_i,a_j)}.
\end{equation}
Each episode of the MCTS proceeds to build a portion of the search tree $\mathcal{T}^{\mathcal{G}' }$ with a depth $n_{\mathcal{G}' }$ and ends the tree policy at a leaf node $\mathcal{N}_l$ in the state $s_l$ with the corresponding subgraph given by $\mathcal{G}_l'$. Here, instead of episode depth $n_{\mathcal{G}' }$, we may alternatively use the episode pruning budget $\mathcal{B}_{\mathcal{G}' }=|(\mathcal{G}_l' )^i |-|(\mathcal{G}_l' )^{i+1} |$, i.e., the difference between number of nodes of the leaf nodes at the end of the episode $i$ and $i+1$. This simply means that we remove $\mathcal{B}_{\mathcal G' }$ nodes from the previous episode leaf node in each episode. The size of $\mathcal{B}_{\mathcal{G}' }$ ultimately depends on the complexity and the size of the search space and other hyperparameters associated with the contextual bandit algorithm that allocates global pruning budgets among the three search spaces. For example, one may decide to utilize a variable pruning budget across episodes, i.e., $\mathcal{B}^i_{\mathcal{G}'}$ that can streamline pruning strategy at initial (or ending) episodes. After each episode, where a tree depth $n_{\mathcal{G}' }$ of $\mathcal{T}^{\mathcal{G}' }$ is built, the MCTS parameters are updated as follows:
\begin{equation}
\label{44}
    \mathrm{N}(s_i, a_j)=\mathrm{N}(s_i,a_j)+1,
\end{equation}
\begin{equation}
\label{45}
    w(s_i,a_j)=w(s_i,a_j)+\mathcal{V}(s_l),
\end{equation}
where one can interpret $\mathcal{V}(s_l )$ as an estimate of the value of the leaf node or final reward received after a given episode ends.\footnote{Note in the case of games $\mathcal{V}(s_l )$ measures how good the chance of a player to win the game is at the leaf state (end of an episode).}  Now, the question is how to define $P(s,a)$ and $\mathcal{V}(s_l)$ properly based on game values given in Eq. \eqref{34}. Here, we assume the following choices:
\begin{equation}
\label{46}
    P(s_i, a_j) =\varphi^{(S',\mathcal{G}'_j, M)}(\mathcal{G}'_j),
\end{equation}
\begin{equation}
\label{47}
    \mathcal{V}(s_l)=\varphi^{(S',\mathcal{G}'_l,M)}(\mathcal{G}'_l).
\end{equation}
This choice is inspired by our game theoretic approach discussed in Section \ref{exdws}, which assigns the leaf node payoff to the players based on their fair contributions. One may be tempted to use other scoring functions in formulating the MCTS algorithm such as fidelity scores \citep{mahlau2023re} (see, Eqs. \eqref{311} and \eqref{312}), which are cheaper computationally. However, these approaches fail to consider all the interactions (permutations) between players, which, in turn, results in poor partial explanations. Employing a fundamental game theoretic approach to explain the neural graph embeddings in downstream machine learning tasks (the very problem this paper is trying to address) is even more critical. This is because in a downstream model both tabular features and neural graph embeddings are combined and treated as a new set of feature vectors. A viable explanation approach must be meaningful for explaining both tabular features and graph embeddings when combined together as well as in extreme cases where either the graph embeddings or tabular features are dominant in terms of their importance attributions. Although there are many non-game-theoretic ad hoc approaches available for geometric data (see, e.g., \citep{ying2019gnnexplainer_22,luo2020parameterized_23}), they would not be useful in our context, where both geometric and non-geometric data are concurrently provided to the machine learning pipeline. On the other hand, game-theoretic approaches for explaining machine learning models on non-geometric data are well-adapted. This is the reason we try to adapt a game-theoretic approach for our problem at hand. Hence, we developed \textsc{Mb}\textsc{Explainer}, which offers a unified game-theoretic solution to explain a model prediction when both geometric and non-geometric data are simultaneously present. Finally, the MCTS algorithm for a single episode of the search tree $\mathcal{T}^{\mathcal G'}$ is shown in Algorithm \ref{algo_3}.
\begin{algorithm}
    \SetAlgoLined 
    \KwIn{$\left(\left(S'\right)^i, \left(\mathcal{G}'_l\right)^i,\left(M\right)^i\right)$, $f^e$, $f^d$, $\mathcal{B}_{\mathcal G'}$, $\left(\mathcal{N}_l\right)^i$.}
    \Init{$\mathcal{G}'_{{curNode}}\leftarrow\left(\mathcal{G}'_l\right)^i$. }
    \BlankLine
    \While{$|(\mathcal G'_l)^i|-|(\mathcal{G}'_{{curNode}})|<\mathcal{B}_{\mathcal{G}'}$,}
    {
    \For{all possible pruning actions of $\mathcal{G}'_{{curNode}}$}
    {
        \begin{itemize}
        \item Apply the action $a_j$, i.e., remove a corresponding node from $\mathcal{G}'_{{curNode}}$.
        \item Obtain the child node $\mathcal{N}_j$ in the state $s_j$ with its subgraph $\mathcal{G}'_j$.
        \item Check if the value of $P\left(s_{{curNode}},a_j\right)$ was computed before. If not find $P\left(s_{curNode}, a_j\right)=\varphi^{\left(\left(S'\right)^i,\mathcal{G}'_j,\left(M\right)^i\right)}\left(\mathcal{G}'_{j}\right)$, for the action $a_j$ using Eq. \eqref{46}.
        \item Record the value $P\left(s_{curNode}, a_j\right)$ for possible use in future episodes.
        \end{itemize}
    }
    Using the action selection Eq. \eqref{41}, select the child node $\mathcal{N}_{nextNode}$, update the state $s_{nextState}$ and the search path $curPath=curPath+\mathcal{N}_{nextNode}$. 
    }
    \KwOut{
    \begin{itemize}
        \item Episode ends: Obtain the lead node $(\mathcal{N}_l)^{i+1}$ in the leaf state $s_l$ with the corresponding subgraph given by $(\mathcal{G}'_l)^{i+1}$.
        \item Update the MCST parameters using Eqs. \eqref{44} and \eqref{45}.
    \end{itemize}     
    }
\caption{EPISODIC MCTS FOR THE SUBGRAPH SEARCH TREE  $\mathcal{T}^{\mathcal{G}'}$}\label{algo_3}
\end{algorithm}
One can similarly define the corresponding MCTS algorithms for $\mathcal{T}^{S'}$ and $\mathcal{T}^M$ based on Eqs. \eqref{33} and \eqref{35}, respectively.
\begin{remark}[Comparison with the MCTS in AlphaGo Zero]
    In this remark, we compare the MCTS algorithm we used for the local search spaces with that of the original AlphaGo Zero paper \citep{silver2017mastering_15}. Similar to AlphaGo Zero, in order to approximate the state value, we do not use a default policy to perform a simulation (or roll-out). Instead, we directly use Eqs. \eqref{46} and \eqref{47} with Monte Carlo (MC) sampling \citep{strumbelj2010efficient,yuan2021explainability_13} to achieve an estimate of the values for $\mathcal{V}(s_l )$ and $P(s_i,a_j )$, respectively. Note that AlphaGo Zero uses an indirect approach to determine the values of $\mathcal{V}$ and $P$. In particular, AlphaGo Zero employs a single neural network $f_{\theta}$ that given a state $s$ outputs $(\boldsymbol{P},v) = f_{\theta}(s)$ \citep{CSE599i_21} such that $\boldsymbol{P}$ gives a probability distribution over all possible actions in the state $s$ and $v$ being the value estimate of the state. Although the neural network in AlphaGo Zero is trained using self-played games,\footnote{Note that the previous versions of AphaGo used training data from expert human-played games or a mix of both self-played and human-played Alpha games.} we believe building such a network for an explainer is not appropriate. For one thing, a GNN explainer pruning space does not possess the particular structures of a board game such as AlphaGo Zero. Moreover, we want the explainer not to depend on training other neural networks and directly generate explanations given the trained downstream and graph embedding models. Finally, generating a training dataset for an explainer is a difficult task. This makes it especially difficult to make the explainer algorithm compatible for all the different machine learning tasks on graphs since it needs task-specific data for training the mentioned neural network.
\end{remark}
\begin{remark}[Insights about the MCTS parallelization for Algorithm \ref{algo_3}]
    Note that the for-loop in Algorithm \ref{algo_3}, i.e., computing action values for all possible actions of a root node can be easily parallelized across multiple central processing units (CPUs) or graphical processing units (GPUs). In general, the common parallelization methods for the MCTS \citep{chaslot2008parallel_29,bourki2011scalability_30,yang2020practical_31,liu2020effective_32} can be tailored and applied to the pipeline. 
\end{remark}
\begin{remark}[further speedup of Algorithm \ref{algo_3} using the Monte Carlo (MC) Sampling]
    Note that for further speedup of Algorithm \ref{algo_3}, we use the MC Sampling (see, e.g., \citep{kotsiopoulos2023approximation_33,mitchell2022sampling_34,ghorbani2019data_35,okhrati2021multilinear_36}). In particular, consider the equation for computing $P\left(s_i,a_j \right)=\varphi^{(S',\mathcal{G}'_j,M) } (\mathcal{G}'_j )$, we define the outline of the following MC Sampling algorithm (see, also \citep{yuan2021explainability_13}):
    \begin{itemize}
        \item Sample a random coalition set $S_2$ from $P^{\mathcal{G}'_j}\setminus\left\{\left(\mathcal{G}'_j\right)^M\right\}$, where
        \begin{equation*}
        P^{\mathcal{G}'_j}=\left\{\left(\mathcal{G}'_j\right)^M,\overbrace{v_{k+1},\cdots,v_m}^{\text{nodes not in}\, \mathcal{G}'_j}\right\}. 
        \end{equation*}
        Note that for the sake of simplicity, the random coalition is drawn from a particular probability distribution (see, e.g., \citep{vstrumbelj2014explaining,kotsiopoulos2023approximation_33}) such that the weights in the Shapley value calculations disappear (become unit). 
        \item Therefore, one obtains:
        \begin{equation*}
        \varphi^{(S',\mathcal{G}'_j,M)}\left(\mathcal{G}
        '_j\right):=\mathlarger{\sum}_{S_2} 1.\bigg[f^d\left\{x_{S'},x^*_{N\setminus S'},f^e\left(\left(\mathcal{G}'_j\cup S_2\right)^M\right)\right\}-f^d\left\{x_{S'},x^*_{N\setminus S'},f^e\left(\left(S_2\right)^M\right)\right\}\bigg].
        \end{equation*}
        \item Take the average for $T$ Monte-Carlo sampling steps: $\frac{1}{T}\sum_{t=1}^T\varphi^{\left(S',\mathcal{G}'_j,M\right)}\left(\mathcal{G}'_j\right)$.
        \item Similarly, the MC Sampling can be applied to $\varphi^{\left(S',\mathcal{G}',M\right)}\left(S'\right)$ and $\varphi^{\left(S',\mathcal{G}',M\right)}\left(M\right)$.
    \end{itemize}
\end{remark}

\subsubsection{Assembling the global search space}
\label{globalsp}
The next question we need to answer is how we properly allocate pruning budgets among the local search trees, i.e., $\mathcal{T}^{\mathcal G'}$, $\mathcal{T}^{S'}$, and $\mathcal{T}^M$. There are many heuristic approaches that one may adopt, e.g., one can uniformly explore the three local spaces. However, these heuristic approaches are not optimal and often lead to computational inefficiencies, and thus, could result in suboptimal explanations. Therefore, here, we propose to use a contextual bandits type algorithm \citep{cortes2018adapting_25,li2010contextual_26} to efficiently allocate pruning budgets to each search tree. The contextual bandits algorithm gives us the flexibility to utilize various exploration-exploitation strategies depending on possible constraints on the desired explanation. Moreover, a contextual bandits method facilitates experimenting with and choosing different classes of oracles to fit to context-reward pair for each arm (or action). 

\begin{figure}[!htb]
	\centering
    \centerline{\includesvg[width=0.8\columnwidth]{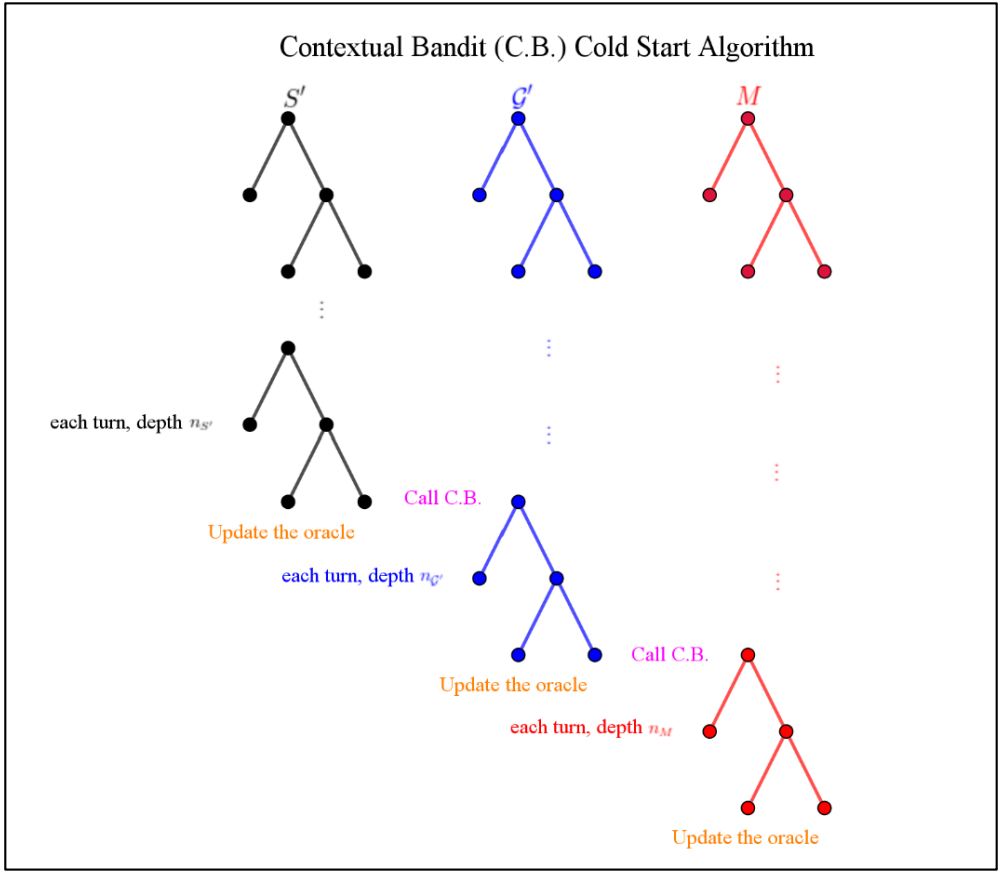}}
	\caption{Assembling the global search space using a contextual bandits algorithm.}
	\label{fig_4}
\end{figure}

Next, we discuss how the bandit algorithm is formulated (see Figure \ref{fig_4}). In doing so, we assume that the arms or actions are the choice of selecting one of the three local search spaces on each episode from $\mathcal{T}^{\mathcal{G}'}$, $\mathcal{T}^{S'}$, and $\mathcal{T}^M$, and proceeding to build their search trees for a fixed incremental depth of $n_{\mathcal{G}'}$, $n_{S'}$, and $n_M$, respectively. Alternatively, we may define an episodic pruning budget (instead of an incremental depth) for $\mathcal{T}^{\mathcal{G}'}$, $\mathcal{T}^{S'}$, and $\mathcal{T}^M$, respectively, as $\mathcal{B}_{\mathcal{G}'}$, $\mathcal{B}_{S'}$, and $\mathcal{B}_M$. The bandit algorithm is initiated with a warm-starting strategy \citep{cortes2018adapting_25,zhang2019warm_27} to address fitting oracles to scenarios with minimal data (or starting from zero, i.e., cold-start problem). At the end of each episode $i$, the contextual bandit algorithm policy $\Pi$ receives the context $\left(S',\mathcal{G}',M\right)$ and the corresponding reward at the end of the episode, adds the event (or observation) to the history of the selected arm, updates the oracle with the new history, and finally, decides that which of the three local search spaces should be pruned next. The details of the contextual bandit algorithm for navigating the global search space are outlined in Algorithm \ref{algo_4}.

\begin{algorithm}
    \SetAlgoLined 
    \KwIn{iterations numbers $\kappa$, root node for $(\mathcal{T}^{\mathcal G' }, \mathcal{T}^{S' }, \textrm{and}\, \mathcal{T}^M)$, episodic pruning budget $\mathcal{B}_{\mathcal G' }$, $\mathcal{B}_{S' }$, and  $\mathcal{B}_M$, oracles  $\hat{f}_{1:3}$ , bandit hyperparameters ($t_b$: breakpoint in Explore-Then-Exploit)}
    \Init{warm-start the contextual bandit algorithm }
    \BlankLine
    \For{$j=1$ to $\kappa$}
    {
        Initialize $\mathcal{T}^{\mathcal G' }$, $\mathcal{T}^{S' }$, and $\mathcal{T}^M$ to their root nodes\\
        \While{$\left(S',\mathcal{G}',M\right)$ does not satisfy \emph{explanation requirements}}
        {
        \For{each successive episode $i$}
        {
            \uIf{$i<t_b$}{
                    Select an action $\mathcal{T}^a$ uniformly at random to choose one of the search trees $\left(\mathcal{T}^{\mathcal{G}'},\mathcal{T}^{S'},\mathcal{T}^M\right)$.
                }            
            \Else{
                Select an action $\mathcal{T}^a=\maxx_{k}\hat{f}_k\left(S',\mathcal{G}',M\right)$.
                }
                \begin{itemize}
                    \item Execute Algorithm \ref{algo_3} to perform pruning of the corresponding local
                    search space.
                    \item Obtain the episode reward $r_a^i$ using \eqref{39} and add the observation $\left\{{(S',\mathcal{G}',M)^i,r_a^i}\right\}$\\ to the history for action $\mathcal{T}^a$. Also, record the calculated reward from \eqref{39} for possible\\ use for future iterations.
                    \item Update the corresponding oracle $\hat{f}_a$ with the new history (or choose a batch size for \\oracle refitting).
                \end{itemize}
        }
        }
        Record the explanation $\left(S',\mathcal{G}',M\right)^j$.
    }
    \KwOut{Select the explanation with the highest score, i.e., $\maxx_{j=1:\kappa}\varphi^{\left(S',\mathcal{G}',M\right)}\left(S',\mathcal{G}',M\right)^j$.}
\caption{EPISODIC CONTEXTUAL BANDITS FOR ASSEMBLING THE GLOBAL SEARCH SPACE}\label{algo_4}
\end{algorithm}
\begin{remark}[explanation requirements and size]
One can enforce various requirements on the explanation $\left(S',\mathcal{G}',M\right)$. It is desired to find the most expressive explanations which are also succinct. Thus, one needs to limit the size of the explanation. One can choose to limit the size of each component of the explanation, i.e.,\footnote{Note that $M^c$ and ${S'}^c$, respectively, denote the node features not masked by $M$ and the downstream features not masked by $S'$.}  
\begin{equation}
\label{49}
    |M^c|\leq N^{\min}_M,\qquad |{S'}^c|\leq N^{\min}_{S'},\qquad |\mathcal{G}'|\leq N^{\min}_{\mathcal{G}'},
\end{equation}
or, alternatively, enforce a limit on the total size of the explanation such that:
\begin{equation}
\label{410}
    |\left(S',\mathcal{G}',M\right)|:=|M^c|+\left|{S'}^c\right|+\left|\mathcal{G}'\right|\leq N^{\min}_{\left(S',\mathcal{G}',M\right)}.
\end{equation}
It is also possible to have a requirement consisting of both Eqs. \eqref{49} and \eqref{410}. One may decide to change the contextual bandits algorithm and tailor it to accommodate the explanation requirement naturally using the most appropriate bandit algorithm. For example, we can devise a budget-limited contextual bandits \citep{tran2012budget_28} with the total budget given by
\begin{equation}
    N^o_{\left(N,\mathcal{G},N\right)}-N^{\min}_{\left(S',\mathcal{G}',M\right)},
\end{equation}
where $N^o$ is the total size of the computational graph initially, i.e., without any node pruning of the subgraph or any masking of the graph features or augmented downstream features. In this case, one can assign arm pulling costs (i.e., the cost or a budget allocated for a particular action selection) relative to the sizes of the pruning spaces as follows:
\begin{equation}
    c_M=\frac{|M^c|+\left|{S'}^c\right|+\left|\mathcal{G}'\right|}{|M^c|}.c_o,\quad c_{S'}=\frac{|M^c|+\left|{S'}^c\right|+\left|\mathcal{G}'\right|}{\left|{S'}^c\right|}.c_o,\quad c_{\mathcal{G}'}=\frac{|M^c|+\left|{S'}^c\right|+\left|\mathcal{G}'\right|}{\left|\mathcal{G}'\right|}.c_o.
\end{equation}
This will encourage the bandit algorithm (and, thus, the global search strategy) to assign higher (lower) costs to components of the explanation with smaller (larger) sizes.
\end{remark}

\begin{remark}[Choice of Multi-Armed Bandits (MAB) vs Contextual Bandits]
    Note that one does not necessarily need to utilize a contextual bandit algorithm to allocate pruning budgets to local search spaces. Contextual bandit leverages the context to make a more informed decision as to which action would have a higher expected reward. However, this comes at the cost of fitting oracles and keeping track of the context-reward pairs. Depending on the use case one can alternatively choose to use an MAB algorithm instead of contextual bandit. The trade-off here is that an MAB type algorithm would need substantially greater number of iterations because it is completely blind to the context. 
\end{remark}
\begin{remark}[Choice of Rewards for Contextual Bandits]
    One may choose to utilize another reward function instead of $\varphi^{(S',\mathcal{G}',M) } \left(S',\mathcal{G}',M\right)$ for the contextual bandits algorithm. One alternative is the Fidelity score (see evaluation framework in Section \ref{dwstream}). In this context, $\text{Fidelity}_+ \left(S',\mathcal{G}',M\right)$ can be used as the reward function, or, alternatively, $\text{Fidelity}_- \left(S',\mathcal{G}',M\right)$ can be used as the regret function. One should note that the Fidelity score is not a game-theoretic type of score. However, a potential issue with using a non-game-theoretic reward function for the global search space pruning is less severe than using a non-game-theoretic reward function for the local search spaces pruning. After all, the contextual bandits (navigating the global search space) is solely responsible for efficiently allocating the total pruning budget to the three local search trees. Therefore, one may choose to utilize a Fidelity score (which is much faster to compute) for the contextual bandit algorithm reward function, and still use the game-theoretic reward function in the MCTS for the local search spaces pruning.
\end{remark}
\begin{remark}[Choice of Contextual Bandits Algorithm and Alternatives]
    Note that in Algorithm \ref{algo_4}, we used one of the simplest bandit strategies, i.e., explore-then-exploit for illustration purposes. We may alternatively use other more sophisticated bandit strategies depending on the computational requirements and the complexity of pruning the local search spaces.
\end{remark}

\subsection{Implementation details for MBExplainer}
\label{impdetails}
In this section, we describe how our methodology is implemented. At a high-level, the algorithm starts out with an explanation $T'=(S',\mathcal{G}', M')$ consisting of the original model input or the output of a previous round of pruning. Next, a pruning method $a$ is picked (randomly or using an oracle) and pruning actions are performed until the budget $\mathcal{B}_a$ is exhausted. Finally, an intermittent reward is computed using the pruned $T'$ which is then propagated to its parents. Then, this process of action selection and pruning is repeated until $T'$ satisfies our explanation requirements. We then repeat this process until we have generated a satisfactory number of explanation candidates and select the best one according to our importance score. For a more detailed outline of this procedure, see Algorithm \ref{algo_impl}. For a demonstration of how the algorithm progresses, see a sample of how \textsc{Mb}\textsc{Explainer} logs its progress in Appendix \ref{MBlog}.

Algorithm \ref{algo_impl} may be directly applied to graph classification models and naturally generalizes to node classification contexts. For example, given that we have an upstream neural graph embedding model $f^e: \mathcal{G}^M \to (\mathbb{R}^D)^{|\mathcal{G}|}$, a downstream model $f^d: \mathbb{R}^N \times \mathbb{R}^D \to \mathbb{R}$ and a prediction for node $v$ that we seek to explain, we may construct a function that outputs the embeddings only for node $v$, $f_{v}^e(\mathcal{G}^M) = f^e(\mathcal{G}_v^M)$ so that the downstream model score $f^d\left(\boldsymbol{x}_v||f^e_v\left(\mathcal{G}_v^M\right)\right)$ may be constructed analogously to how it is done in graph classification. Furthermore, we may reduce the size of $\mathcal{G}$ to just the computational graph of $f^e$ around node $v$ to increase calculation speed without loss of accuracy. Finally, we may proceed to Algorithm \ref{algo_impl} using $f_{v}^e$ as the upstream model and $f^d$ as the downstream model.

With respect to neutralizing nodes, \textsc{Mb}\textsc{Explainer} follows a zero-padding strategy, i.e., the convention of pruning a node by replacing its node features by a zero-vector (see, also \citep{yuan2021explainability_13}). A similar approach is used to prune upstream node features, with each pruned feature being replaced by zero in all nodes of the graph. Downstream features are pruned by replacing their values with their average feature values across the entire downstream (classification task) dataset.

\begin{remark}
Due to \textsc{Mb}\textsc{Explainer} using a zero-padding strategy, it is crucial that the node features are designed in such a way that replacing a feature with zero may be viewed as a sensible way to neutralizing it. This may be the case when every node feature has a mean of zero and may be accomplished through common data standardization techniques. In this work, we employ standard scaling which ensures all upstream features have a mean of zero and standard deviation of one.
\end{remark}

\begin{remark}
One consequence of using a tree policy of the form Eq. \eqref{41} is that all child nodes are initially tied at $Q(s_i, a_j) + U(s_i, a_j) = 0$. This occurs because $Q(s_i, a_j)$ is conventionally set to zero when otherwise undefined. Simultaneously, $U(s_i, a_j)$ is zero since  no children have been visited, and, therefore, $\sum \textrm{N}(s_i, a_j)=0$. In this case, the algorithm proceeds according to how the particular implementation of $\maxx$ breaks ties. Commonly, $\maxx$ breaks ties based on the order in which the potential pruning actions are considered. This may introduce a subtle bias into how the MCTS initially proceeds. For \textsc{Mb}\textsc{Explainer}, this bias is exacerbated by the larger search space which means that a particular parent node will be populated with children more slowly. 
In some cases such as node-pruning, this bias may be desirable as it favors pruning high degree or low degree nodes (depending on \textsc{Low2high} or \textsc{High2low} setting \citep{yuan2021explainability_13}) all else being equal.\footnote{Note that \textsc{Low2high} and \textsc{High2low} are settings in the implementation of \citep{yuan2021explainability_13} that determine which nodes to consider pruning during the MCTS. Considering all nodes adjacent to the current subgraph may be expensive so node consideration is limited to the top $n_{child}$ nodes in order of least (\textsc{Low2high}) or greatest (\textsc{High2low}) node degree.} However, in other cases which lack a property analogous to node degree, such as feature pruning, the bias may not be beneficial. For example, a naive feature pruning procedure may prune actions in order of what features come first in the dataset. In this case, the MCTS search first prunes feature 1, then feature 2, then feature 3, and so on to yield a final explanation with a consecutive block of feature indices. Datasets often do not order their features in a random way so this may create the misleading impression that the MCTS rollout has identified a pattern.
To eliminate this bias, one may randomize the order in which prospective pruning actions are considered. Alternatively, one may use the prior $P(s_i, a_j)$ as a tie-breaker for actions when none of their resultant child nodes have been visited before. This may be viewed as a variant of the optimism in the face of uncertainty \citep{sutton2018reinforcement} and is shown in Algorithm \ref{algo_impl}.
\end{remark}
 \begin{algorithm}
    \SetAlgoLined 
    \KwIn{iterations numbers $\kappa$, model input tuple ($S$, $\mathcal{G}$, $M$) episodic pruning budgets $\mathcal{B}_{1:3}$, oracles  $\hat{f}_{1:3}$, constraints on explanation sizes $N^{\min}_{1:3}$ for ($S$, $\mathcal{G}$, $M$), bandit hyperparameters ($t_b$: breakpoint in Explore-Then-Exploit, $\alpha$: Explore vs Exploit trade-off)}
    \Init{warm-start the contextual bandit algorithm }
    \BlankLine
    Initialize empty maps sending (node, action) tuples to numbers $P, N, w:  (\mathcal{N}, a)\to \mathbb{R}$\\
    \For{$j=1$ to $\kappa$}
    {
        Re-initialize current node to original model input $T'=(S', \mathcal{G}', M')\leftarrow (S, \mathcal{G}, M)$\\
        Re-initialize action set $A\leftarrow \{0, 1, 2\}$\\
        \While{$T'$ does not satisfy \emph{explanation requirements}}
        {
        \For{each successive episode $i$}
        {
            \uIf{$i<t_b$}{
                    Select a pruning action $a\in A$ uniformly at random
                } 
            \Else{
                Select a pruning action $a:=\maxx_{k\in A}\hat{f}_k\left(S',\mathcal{G}',M'\right)$.
                }
            Start recording episode history $H_0^i := T'$\\
            Initialize budget counter $k := 1$\\
            \While{$|T_a'| > N^{\min}_a$ and $k \leq \mathcal{B}_a$}{
                    Initialize set of children $C=\{\}$\\
                    \For{all considered pruning actions $a_m$ associated with $a$ and applicable to $T'$}{
                        Apply action $a_m$ to obtain the child node $\mathcal{N}_m$\\
                        \uIf{$P(\mathcal{N}_m, a)$ does not exist}{
                            $P(\mathcal{N}_m, a):=\varphi_{a}^{\mathcal{N}_m}({\mathcal{N}_m})$\\
                            $N(\mathcal{N}_m, a):=0.0$\\
                            $w(\mathcal{N}_m, a):=0.0$\\
                        }
                        Update child set $C \leftarrow C \cup {\mathcal{N}_m}$
                    }
                    select child node according to Eq. \eqref{41} $T_{child}' := \maxx_{\mathcal{N} \in C} \frac{w(\mathcal{N}, a)}{\max(N(\mathcal{N}, a), 1)} + \alpha \cdot P(\mathcal{N}, a)\frac{\sqrt{\max(\sum_{\mathcal{N}' \in C} N(\mathcal{N}',a), 1)}}{1+N(\mathcal{N}, a)}$\\
                    Update current node to child node $T' \leftarrow T_{child}'$\\
                    \uIf{$|T_a'| \leq N^{\min}_a$}{
                        $A := A\backslash\{a\}$
                    }
                    Update episode history $H_k^i := T_{child}'$\\
                    Update budget $k \leftarrow k + 1$
            }
            Obtain episodic reward $r^i$ using Eq. \eqref{39}\\
            \For{$\mathcal{N} \in H^i$}{
                Propagate visit counts $\textrm{N}(\mathcal{N}, a)\leftarrow \textrm{N}(\mathcal{N}, a) + 1$\\
                Propagate total reward $w(\mathcal{N}, a)\leftarrow w(\mathcal{N}, a)+r^i$
            }
            Update oracle $\hat{f}_a$ with new observation $\left\{{H_{0}^i,r_a^i}\right\}$. Note that this is only necessary when we expect to need the oracle's predictions before its history is updated again.
        }
        }
        Record the explanation $E_j := T'$.
    }
    \KwOut{Select the explanation in $E$ with the highest score, i.e. using Eq. \eqref{39}}
\caption{\textsc{Mb}\textsc{Explainer} Pseudocode Implementation Details}\label{algo_impl}
\end{algorithm}


\section{Experiments}
\label{numerics}
In this section, through several detailed numerical examples we demonstrate how \textsc{Mb}\textsc{Explainer} can be applied to downstream models with neural graph embeddings. In particular, we study \textsc{Mb}\textsc{Explainer}'s performance for graph classification tasks using the MUTAG, PROTEINS, and ZINC datasets as well as node classification tasks on the ogbn-arxiv dataset. Note that all these datasets were downloaded from the Deep Graph Library (DGL) \citep[v1.1.0]{wang2019dgl}.

\subsection{MUTAG}
\label{mutag}
The MUTAG dataset \citep{doi:10.1021/jm00106a046} contains $188$ nitroaromatic compounds represented as graphs with nodes corresponding to atoms and edges corresponding to bonds. Each compound also has a binary label corresponding to its mutagenicity on Salmonella typhimurium. Each node has a seven dimensional node feature which represents its atom type via one-hot encoding. We further augment each compound in MUTAG with tabular/graph-level features corresponding to the total count it contains of each atom type by summing up its node features. To train a model, we partition MUTAG into $169$ compounds for training and $19$ compounds for validation. We also rescale the one-hot features by removing their mean and dividing by the standard deviation based on the training dataset.

We construct an upstream graph classification model by training a Graph Isomorphism Network (GIN) \citep{xu2018how} on MUTAG to predict compound mutagenicity. All models are GINs with four hidden layers and hidden layer dimension of eight. We use the Adam optimizer \citep{kingma2017adammethodstochasticoptimization} to train for $350$ epochs using a batch size of $128$, and an initial learning rate of $0.01$ that decays by a factor of $\gamma=0.5$ every fifty epochs.

Following this, we retrieve the final hidden layer representation of the trained GIN to produce an eight dimensional graph embeddings. We further concatenate the neural graph embeddings with the total counts of each atom type (seven features) and represent the combined features as a tabular dataset. We employ this dataset to train a downstream CatBoost model ($5$ rounds of early-stopping, default parameters) to predict mutagenicity.

Finally, we may employ \textsc{Mb}\textsc{Explainer} to produce multilevel explanations for our model. In this example, we let $\kappa=20$, $\mathcal{B}_{1:3}=(3, 5, 5)$, $N^{\min}_{1:3}=(2,5,2)$ and estimate Shapley values using $100$ permutations. Furthermore, we weight nodes, graph features and downstream features equally in our importance score (Eq. \eqref{39} with $\lambda^M=\lambda^{\mathcal{G}'}=1$) and use a random bandit policy. For comparison, we also employ SubgraphX in explaining the upstream GIN directly.
\begin{figure}[!htb]
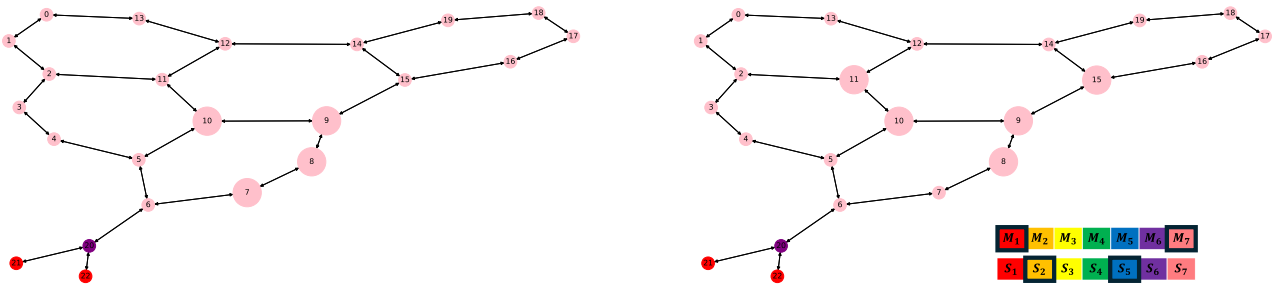

	\centering
    \centerline{    \includesvg[width=0.55\columnwidth]{Figures/mutag_subgraphx.svg}
    \includesvg[width=0.55\columnwidth]{Figures/mutag_mb.svg}
    }
	\caption{Example visualization comparing SubgraphX's explanation of the upstream GIN (left) with \textsc{Mb}\textsc{Explainer}'s explanation of the downstream model (right) on MUTAG. Nodes are large when selected as part of an explanation and colored based on atom type ($M_1$=1, $M_6$=1 and $M_7=1$ nodes are red, pink and purple, respectively). Upstream graph features and downstream features selected by \textsc{Mb}\textsc{Explainer} are emphasized by thick black boxes. Both upstream and downstream features correspond to atom types are colored accordingly.}
    \label{fig_mutag}
\end{figure}

The node explanation produced by \textsc{Mb}\textsc{Explainer} and SubgraphX shown in the Figure \ref{fig_mutag} have a lot of overlap (nodes 8, 9, 10). \textsc{Mb}\textsc{Explainer} also selected two graph feature indices ($1$ and $7$) and two augmented downstream feature indices ($2$ and $5$) as explanatory. Indices $1$ and $7$ correspond to the standardized one-hot encodings of the two atom types present in the example compound. While indices $2$ and $5$ do not correspond to atom types in the example compound, they have the greatest CatBoost feature importance scores of all the downstream features provided to it.

\subsection{PROTEINS}
\label{proteins}
The PROTEINS dataset 
\citep{10.1093/bioinformatics/bti1007} consists of $1113$ proteins where nodes are secondary structure elements (SSEs) of those proteins and edges exist if SSEs are less than six angstroms apart or are neighbors in the amino-acid sequence. Each protein has a binary label corresponding to whether or not it is an enzyme. Each node has a three dimensional node feature which represents its SSE type (helix, sheet or turn) via one-hot encoding. We further augment each protein with tabular/graph-level features corresponding to the total count it contains of each SSE type (three features) by summing up the node features. We randomly partition the dataset into $1001$ compounds for training and $112$ compounds for validation. We also rescale the node features as was done in Section \ref{mutag} for the MUTAG dataset.

We proceed by constructing the upstream and downstream models in a manner identical to the approach with MUTAG \S\ref{mutag} and generate explanations with the same parameters.

\begin{figure}[!htb]
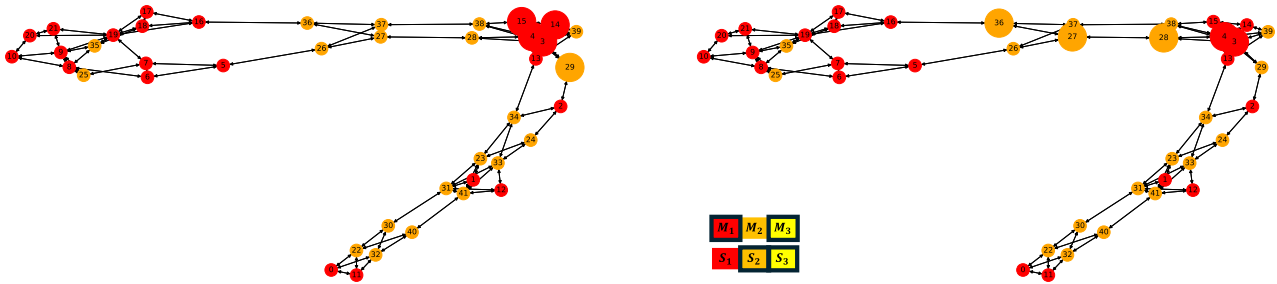

	\centering
    \centerline{
    \includesvg[width=0.55\columnwidth]{Figures/proteins_subgraphx.svg}
    \includesvg[width=0.55\columnwidth]{Figures/proteins_mb.svg}
    }
	\caption{Example visualization comparing SubgraphX's explanation of the upstream GIN (left) with \textsc{Mb}\textsc{Explainer}'s explanation of the downstream model (right) on PROTEINS. Nodes are large when selected as part of an explanation and colored based on atom type ($M_1$=1 nodes are red and $M_2$=1 nodes are orange). Upstream graph features and downstream features selected by \textsc{Mb}\textsc{Explainer} are emphasized by thick black boxes. Both upstream and downstream features correspond to atom types are colored accordingly.}
    \label{fig_proteins}
\end{figure}

 The node explanations for an example graph in Figure \ref{fig_proteins} contain some overlap (node $36$) but are structurally distinct. For this example, \textsc{Mb}\textsc{Explainer} also selects two upstream graph feature indices ($1$ and $2$) and two augmented downstream feature indices ($2$ and $3$). Feature indices $1$ and $2$ correspond to atom types that are present in the molecule while index $3$ does not. Nevertheless, feature index 3 is considered the most important downstream feature by the CatBoost model. The second most important feature index according to the CatBoost model is $1$. However, its presence in the upstream feature portion of the explanation may render its inclusion unnecessary in the downstream portion.

\subsection{ZINC (Binary Classification)}
\label{zinc}
The subset of the ZINC dataset we employ \citep{10.5555/3648699.3648742} consists of $12,000$ molecules represented with nodes corresponding to atoms and edges corresponding to bonds. From these $12,000$ molecules, $1,000$ of them are reserved for a validation and test set each. Each molecule has a label which quantifies its constrained solubility. Each node and bond have one-dimensional features describing what type of atom and what type of bond they correspond to, respectively. To prepare the dataset for graph classification, we one-hot encode both the node feature and edge feature to produce a $21$ dimensional node feature and three dimensional edge feature. We also rescale the node features using the training and validation set as was done in the MUTAG experiment. Finally, we create a binarized label which is one if the constrained solubility exceeds its median value of about $0.4407$ on the training dataset.

We further augment each molecule with seven tabular/graph-level features engineered from the one-hot encoded edge features. These engineered features include the total count of each bond type in the molecule, the proportion of bonds made-up of each bond type, and the atom-to-bond count ratio. We construct the upstream and downstream models using the same methodology as in the MUTAG experiment. Note that our GIN upstream model does not make use of edge features so only our downstream CatBoost model may make use of them, via our engineered augmented downstream features.
\begin{figure}[!htb]
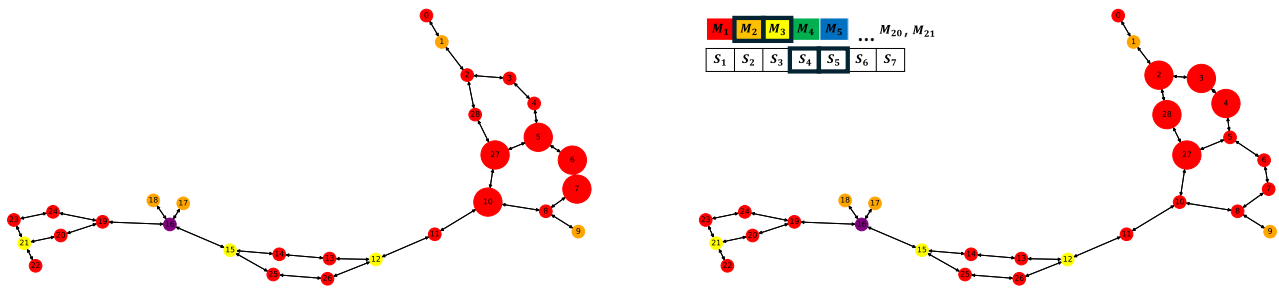

	\centering
    \centerline{
    \includesvg[width=0.55\columnwidth]{Figures/zinc_subgraphx.svg}
    \includesvg[width=0.55\columnwidth]{Figures/zinc_mb.svg}
    }
	\caption{Example visualization comparing SubgraphX's explanation of the upstream GIN (left) with \textsc{Mb}\textsc{Explainer}'s explanation of the downstream model (right) on ZINC. Nodes are large when selected as part of an explanation and colored based on atom type ($M_1$=1 nodes, $M_2=1$, $M_3=1$, and $M_6$=1 are red, orange, yellow, and purple, respectively). Upstream graph features and downstream features selected by \textsc{Mb}\textsc{Explainer} are emphasized by thick black boxes. Both upstream and downstream features correspond to atom types are colored accordingly.}
    \label{fig_zinc}
\end{figure}

On the example graph shown in Figure \ref{fig_zinc}, SubgraphX and \textsc{Mb}\textsc{Explainer} identify overlapping subgraphs, but with distinct structures in the molecule. Both of these structures appear to have the same 2-hop computational graph, consisting of chains of five $M_1=1$ atoms that are one atom away from being cyclic. For this example, \textsc{Mb}\textsc{Explainer} also selects two upstream features ($2$ and $3$) and two augmented downstream feature indices ($4$ and $5$). The upstream features correspond to atom types $2$ and $3$ which are present in the molecule. The downstream feature indices $4$ and $5$ are the third and first most important downstream features according to CatBoost and correspond to the percentage of type $1$ and $2$ bonds in the molecule.

\subsection{ogbn-arXiv}
The ogbn-arXiv dataset consists of a 90941 node training set, 29799 node validation set and 48603 node test dataset where each node corresponds to an arXiv paper \citep{10.5555/3495724.3497579}. Each paper also corresponds to one of forty arXiv subject areas. While ogbn-arXiv edges are directed and correspond to when one paper cites another, we append reverse edges for the purpose of link prediction. Each arXiv paper node is associated with a 128 dimensional feature vector obtained by averaging embeddings of words in its title and abstract.

For the upstream model, we construct a two layer GraphSAGE neural network with 64 dimensional hidden layers and a 32 dimensional output layer. It learns to produce node embeddings such that the dot product between the embeddings of two nodes corresponds to the logit of the probability that a link exists between them. Uniform neighborhood sampling is employed with four 1-hop and four 2-hop neighbors being sampled. The model is trained for one epoch.

We use this upstream model to generate node embeddings for the whole ogbn-arXiv dataset and generate a tabular dataset combining these node embeddings with the original word embedding features. Finally, we train a CatBoost model (100 epochs iterations, default parameters) on this tabular dataset to predict its subject area.

\begin{figure}[!htb]
	\centering
    \centerline{
    \includesvg[width=0.85\columnwidth]{Figures/ogbn_mb_25.svg}
    }
	\caption{Example visualization of \textsc{Mb}\textsc{Explainer}'s explanation of the downstream model for ogbn-arXiv. Nodes are colored based on paper subject area. Circular nodes correspond to nodes in the explanation while the large square node corresponds to the node whose score is being explained. Thick solid black edges, thin solid black edges, thin gray edges and thin translucent edges correspond to edges between explanation nodes, edges between explanation nodes and the node being explained, edges connected to an explanation node, and all other nodes, respectively. Upstream graph features and augmented downstream features selected by \textsc{Mb}\textsc{Explainer} are shown as well.}
    \label{fig_ogbn}
\end{figure}

\begin{figure}[!htb]
	\centering
    \centerline{
    \includesvg[width=1.05\columnwidth]{Figures/least_most_combined_ogbn.svg}
    }
	\caption{Example visualization of \textsc{Mb}\textsc{Explainer}'s explanation of the downstream model for ogbn-arXiv. Nodes are colored based on paper subject area. Circular nodes correspond to nodes in the explanation while the large square node corresponds to the node whose score is being explained. Thick solid black edges, thin solid black edges, thin gray edges and thin translucent edges correspond to edges between explanation nodes, edges between explanation nodes and the node being explained, edges connected to an explanation node, and all other nodes, respectively. Upstream graph features and augmented downstream features selected by \textsc{Mb}\textsc{Explainer} are shown as well.}
    \label{fig_ogbn2}
\end{figure}

Finally, we may employ \textsc{Mb}\textsc{Explainer} to produce multilevel explanations for our model. In this example, we adjust the budget and explanation requirements to accommodate the larger graph size and feature space. We let $\kappa=1$, $\mathcal{B}_{1:3} = (6, 18, 6)$, $N^{\min}_{1:3} = (6, 18, 6)$ and estimate Shapley values using $25$ permutations. Figure \ref{fig_ogbn} depicts the results. The node being explained is included in the explanation and connected directly to five other explanation nodes, two of which are highly central to the computational graph. Features 13 (selected as an upstream graph feature), 4 (selected as an upstream graph feature), and 20 (selected as a downstream feature) correspond to the first, third, and fourth most important downstream features, respectively (excluding graph embeddings), based on the built-in downstream CatBoost feature importance.

Figure \ref{fig_ogbn2} shows additional explanations computed for ogbn-arXiv, one for a node with the least common label and one for a node with the most common label. We observe similar patterns. The explanation for the node with the least common label includes the node being explained along with features five and 124. These correspond to the fifth and second most important features, respectively, according to CatBoost. For the node with the most common label, the explanation does not include the node being explained but does include feature 124 again, along with feature 13, which is the most important downstream feature according to CatBoost.

\subsection{Results Summary}
\label{num_sum}
\begin{table}
\centering
\begin{tabular}{@{}lllll@{}}
\toprule
 & \multicolumn{4}{c}{Dataset} \\ \midrule
 & MUTAG & PROTEINS & ZINC & ogbn-arXiv \\
 \midrule
\# of Edges (avg) & 39.59 & 145.63 & 49.83 & 1166243 \\
\# of Nodes (avg) & 17.93 & 39.06 & 23.15 & 169343 \\
\# of Node Features & 7 & 3 & 21 & 128 \\
\# of Edge Features & NA & NA & 3 & NA \\
\# of Graphs & 188 & 1113 & 12000 & 1 \\
\# of Classes & 2 & 2 & 19 & 40 \\ \bottomrule
\end{tabular}
\caption{High-level statistics and properties of datasets.}
\label{dataset_properties}
\end{table}

Table \ref{dataset_properties} describes the properties of these datasets at a high level. Table \ref{experiment_table} summarizes certain quantitative observations about the experiment. On MUTAG and PROTEINS, \textsc{Mb}\textsc{Explainer} is somewhat slower than SubgraphX (given the extra work it needs to do) but much slower than SubgraphX on ZINC. This is because pruning nodes in the ZINC example may be accomplished rapidly  (due to low connectivity of nodes) while pruning features requires more time (due to the large number of features). As a result, both SubgraphX and \textsc{Mb}\textsc{Explainer} find the node component of the explanation quickly, but \textsc{Mb}\textsc{Explainer} allocates a considerable amount of time for pruning node features.

\begin{table}[]
\centering
\begin{tabular}{@{}llllllll@{}}
\toprule
\multicolumn{4}{l}{Experiment} & 
\multicolumn{2}{l}{Accuracy} & \multicolumn{2}{l}{Duration (min)} \\ \midrule
Dataset & $|\mathcal{G}|$ & $|M|$ & $|S|$ & GNN & Downstream & SubgraphX & \textsc{Mb}\textsc{Explainer} \\ \midrule
MUTAG & 23 & 7 & 7 & 0.8947 & 0.8421 & 7 & 14 \\
PROTEINS & 42 & 3 & 3 & 0.7411 & 0.7143 & 21 & 28 \\
ZINC & 29 & 21 & 7 & 0.9130 & 0.9100 & 4 & 42 \\
ogbn-arXiv & 367, 17, 601 & 128 & 128 & NA & 0.5211 & NA & 282, 75, 350 

\\ \bottomrule
\end{tabular}
\caption{Table summarizing aspects of explainability experiments. For MUTAG and PROTEINS, performance shown is for the validation dataset. For ZINC and ogbn-arXiv, performance shown is for the test dataset. The time duration reported are for single instance predictions. $|\mathcal{G}|$ and the time duration for ogbn-arXiv correspond to the explanations in Fig. \ref{fig_ogbn}, Fig. \ref{fig_ogbn2}-left, and Fig. \ref{fig_ogbn2}-right, respectively.}
\label{experiment_table}
\end{table}

\begin{remark}
One should note that the results for SubgraphX on graph classification tasks are included only as a reference. \textsc{Mb}\textsc{Explainer} generates explanations for the whole pipeline which includes both the downstream model, the GNNs, their respective components and their interactions. On the other hand, SubgraphX only generates explanations for the predictions based on the GNNs and is not accounting for the downstream model. Note also that the SubgraphX implementation that we have used can only explain GNNs on graph classification tasks. That is why we do not report the SubgraphX performance on the ogbn-arxiv dataset which is used for node classification tasks.   
\end{remark}

\begin{remark}
As discussed earlier in \S\ref{dwstream}, treating the graph embeddings as synthetic set of features and applying self-interpretable explainers in the downstream, one inevitably ignores the dependencies of the embeddings and the augmented downstream features. The CatBoost feature importance methodology used here does not depend on upstream nodes or node features, and, therefore, cannot be used to explain all components of the model and their interactions in the explanation. However, we analyze them here merely for the purpose of comparison with the results of \textsc{Mb}\textsc{Explainer}'s explanations.
\end{remark}


\section{Conclusions}
In many practical applications, graph embeddings generated from a trained GNNs model in the upstream are augmented with additional tabular features and utilized in order to train a downstream model for a particular prediction task. As a matter of fact, when it comes to encoding geometric data, GNNs are very efficient, but they often demonstrate suboptimal performance on tabular data. On the other hand, tree-based models such as GBDTs and CatBoost perform very well on tabular data and can natively encode categorical features and handle missing values. Therefore, it is very common to encounter downstream models with augmented neural graph embeddings, making the development of explainability approaches for these ensemble models critically important. Although there are existing interpretability methods for both GNNs and tabular ML models, they are not sufficient to fully explain these downstream models. For instance, employing tabular post-hoc (or self-interpretable) explainers for the downstream model, one would inevitably treat graph embeddings as extra synthetic features (which is equivalent to freezing the components of the upstream computational graph). This, in turn, would result in overlooking the dependencies and complex interactions of neural graph embeddings and augmented tabular features, leading to obtaining incomplete and misleading explanations.  

In this paper, we proposed \textsc{Mb}\textsc{Explainer}, an interpretability approach to comprehensively explain downstream models with augmented graph embeddings in a single pipeline. To the best of our knowledge, this is the first attempt in generating explanations for pipelines that consist of two fundamentally distinct components, namely: graph embeddings with their complex geometric structure and augmented tabular features in downstream having no geometric structure. \textsc{Mb}\textsc{Explainer} returns a human-legible triple consisting of the most important subgraphs, their node features, and augmented downstream features as explanations. The method enjoys a game-theoretic formulation to account for the contributions of each component. In particular, three Shapley values are assigned each corresponding to their specific games in the spaces of downstream features, subgraphs, and node features. The formulation also considers the complex interactions between these spaces when corresponding Shapley values are calculated.   

To ensure the computational cost of generating an explanation remains manageable, \textsc{Mb}\textsc{Explainer} enjoys a novel multilevel searching strategy that makes it possible to simultaneously prune the spaces of subgraphs, their node features, and downstream features as the local search spaces. The local search algorithm is navigated using three interweaved Monte Carlo Tree Search which is responsible for iterative pruning of the local search spaces. The global search strategy leverages contextual bandits algorithm which efficiently allocates pruning budget to drive search among the three local search spaces.

\textsc{Mb}\textsc{Explainer} is applicable to a broad range of machine learning prediction tasks on graphs and is model-agnostic in terms of both the upstream GNNs and the tabular ML model in the downstream. To demonstrate the effectiveness of our method, we presented a series of numerical experiments on public datasets. In particular, we presented \textsc{Mb}\textsc{Explainer}'s performance on the MUTAG, PROTEINS, and Binarized ZINC for graph classification tasks and on the ogbn-arxiv dataset for node classification tasks. 
\section*{Acknowledgement}

We would like to express our gratitude to Sharon O'Shea Greenbach (Sr. Counsel \& Director, Regulatory Policy at DFS) for her helpful comments relevant to
regulatory issues that arise in the banking industry, and for carefully reviewing the manuscript. Ashkan Golgoon benefited from stimulating discussions with Matthias Fey and Rex (Zhitao) Ying. The views and opinions expressed in this paper are solely those of the authors and do not necessarily reflect the official policy or position of our employer. Any content provided is of the authors' own research and perspective and is not intended to malign any organization, company, or individual.

\section{Appendix}
\subsection{Example MBExplainer Log Output}
\label{MBlog}
\textsc{Mb}\textsc{Explainer} is implemented with an optional logging functionality which records how the pruning of an explanation progresses. An example log is shown below, corresponding to the first \textsc{Mb}\textsc{Explainer} iteration in our PROTEINS experiment \S\ref{proteins}.

The log begins by announcing the start of an MCTS rollout, describing how many rollouts have been performed and the number of subgraphs that have been explored. At the start of this rollout, we return to an initial explanation composing the whole graph and all upstream and downstream features. The log then continues with the next step, determining the appropriate pruning action to apply to our starting explanation. Until a user-specified number of rollouts are completed (in this case, five), this pruning action is determined randomly. Afterward, it is determined by user-specified oracles.

With an initial explanation defined and a pruning action defined, the log continues by describing the progress of the pruning procedure. The log reports some details of the initial explanation, along with some details of the child explanation selected using criteria in Eq. \eqref{41}. In this case, the pruning action is being applied to downstream features so the child has one fewer downstream feature than its parent. It turns out that this satisfies the explanation's requirement on the downstream feature count so the log announces that downstream pruning is complete and removes this action from consideration until the end of the rollout. The intermittent reward for this pruning action is recorded by the oracle, but, as the log reports, it is not refit. The oracle is not needed yet so a refit is unnecessary.

With the downstream pruning action episode complete, the log announces that a new pruning action has been selected for a new episode. In this case, nodes are being pruned. As before, details about each parent and selected child node are reported. The parent starts with 42 nodes and a child with 41 nodes is selected to become the new parent. Its selected child has only 34 nodes, likely because removing the pruned node also disconnected a big part of the parent graph (the subgraph explanation must be a connected graph). With each child node selected, the spent budget reported by the log increases by one. After the spent budget reaches the budget of $5$, this new episode ends. As before, the oracle history is updated but the oracle is not refit, and a new action is randomly selected.

This process continues until the end of the log, when the explanation requirement on the graph feature is met and both node and downstream feature explanation requirements have already been met. Since a viable explanation has been generated, this concludes the MCTS rollout.

\begin{verbatim}
Rollout 0/20,                     0 subgraphs have been explored.
selecting game 'downstream' randomly (rollout 0 <= 5)

playing game 'downstream', budget 1/3
parent: 42 nodes, 3 graph features, 3 downstream features
child: 42 nodes, 3 graph features, 2 downstream features

selected child node satisfies downstream count requirement, stopping downstream pruning
Budget ended with 42 nodes, 3 graph features, and with 2 downstream features
Oracle for game 'downstream' is not refit

selecting game 'nodes' randomly (rollout 0 <= 5)

playing game 'nodes', budget 1/5
parent: 42 nodes, 3 graph features, 2 downstream features
child: 41 nodes, 3 graph features, 2 downstream features

playing game 'nodes', budget 2/5
parent: 41 nodes, 3 graph features, 2 downstream features
child: 34 nodes, 3 graph features, 2 downstream features

playing game 'nodes', budget 3/5
parent: 34 nodes, 3 graph features, 2 downstream features
child: 33 nodes, 3 graph features, 2 downstream features

playing game 'nodes', budget 4/5
parent: 33 nodes, 3 graph features, 2 downstream features
child: 32 nodes, 3 graph features, 2 downstream features

playing game 'nodes', budget 5/5
parent: 32 nodes, 3 graph features, 2 downstream features
child: 31 nodes, 3 graph features, 2 downstream features

Budget ended with 31 nodes, 3 graph features, and with 2 downstream features
Oracle for game 'nodes' is not refit

selecting game 'nodes' randomly (rollout 0 <= 5)

playing game 'nodes', budget 1/5
parent: 31 nodes, 3 graph features, 2 downstream features
child: 30 nodes, 3 graph features, 2 downstream features

playing game 'nodes', budget 2/5
parent: 30 nodes, 3 graph features, 2 downstream features
child: 29 nodes, 3 graph features, 2 downstream features

playing game 'nodes', budget 3/5
parent: 29 nodes, 3 graph features, 2 downstream features
child: 28 nodes, 3 graph features, 2 downstream features

playing game 'nodes', budget 4/5
parent: 28 nodes, 3 graph features, 2 downstream features
child: 22 nodes, 3 graph features, 2 downstream features

playing game 'nodes', budget 5/5
parent: 22 nodes, 3 graph features, 2 downstream features
child: 21 nodes, 3 graph features, 2 downstream features

Budget ended with 21 nodes, 3 graph features, and with 2 downstream features
Oracle for game 'nodes' is not refit

selecting game 'nodes' randomly (rollout 0 <= 5)

playing game 'nodes', budget 1/5
parent: 21 nodes, 3 graph features, 2 downstream features
child: 20 nodes, 3 graph features, 2 downstream features

playing game 'nodes', budget 2/5
parent: 20 nodes, 3 graph features, 2 downstream features
child: 19 nodes, 3 graph features, 2 downstream features

playing game 'nodes', budget 3/5
parent: 19 nodes, 3 graph features, 2 downstream features
child: 18 nodes, 3 graph features, 2 downstream features

playing game 'nodes', budget 4/5
parent: 18 nodes, 3 graph features, 2 downstream features
child: 17 nodes, 3 graph features, 2 downstream features

playing game 'nodes', budget 5/5
parent: 17 nodes, 3 graph features, 2 downstream features
child: 16 nodes, 3 graph features, 2 downstream features

Budget ended with 16 nodes, 3 graph features, and with 2 downstream features
Oracle for game 'nodes' is not refit

selecting game 'nodes' randomly (rollout 0 <= 5)

playing game 'nodes', budget 1/5
parent: 16 nodes, 3 graph features, 2 downstream features
child: 9 nodes, 3 graph features, 2 downstream features

playing game 'nodes', budget 2/5
parent: 9 nodes, 3 graph features, 2 downstream features
child: 7 nodes, 3 graph features, 2 downstream features

playing game 'nodes', budget 3/5
parent: 7 nodes, 3 graph features, 2 downstream features
child: 6 nodes, 3 graph features, 2 downstream features

playing game 'nodes', budget 4/5
parent: 6 nodes, 3 graph features, 2 downstream features
child: 5 nodes, 3 graph features, 2 downstream features

selected child node satisfies nodes count requirement, stopping nodes pruning
Budget ended with 5 nodes, 3 graph features, and with 2 downstream features
Oracle for game 'nodes' is not refit

selecting game 'graph_feat' randomly (rollout 0 <= 5)

playing game 'graph_feat', budget 1/5
parent: 5 nodes, 3 graph features, 2 downstream features
child: 5 nodes, 2 graph features, 2 downstream features

selected child node satisfies graph_feat count requirement, stopping graph_feat pruning
Budget ended with 5 nodes, 2 graph features, and with 2 downstream features
Oracle for game 'graph_feat' is not refit
\end{verbatim}

\bibliographystyle{abbrvnat}
\bibliography{ref}





\end{document}